\documentclass{article}

\usepackage[preprint]{neurips_2024}

\usepackage[utf8]{inputenc} 
\usepackage[T1]{fontenc}    
\usepackage{hyperref}       
\usepackage{url}            
\usepackage{booktabs}       
\usepackage{amsfonts}       
\usepackage{nicefrac}       
\usepackage{microtype}      
\usepackage{xcolor}         
\usepackage{chemformula} 
\usepackage{multirow}
\usepackage{subfigure}
\usepackage{rotating}
\usepackage{cleveref}
\usepackage{colortbl}
\usepackage{wrapfig}
\usepackage{threeparttable}

\definecolor{red}{RGB}{192,0,0}
\definecolor{green}{RGB}{84,130,53}
\definecolor{yellow}{RGB}{197,90,17}
\definecolor{grey}{rgb}{0.9,0.9,0.9}

\title{Developing ChemDFM as a Large Language Foundation Model for Chemistry}

\author{%
  Zihan Zhao$^1$\:\!\thanks{Zihan Zhao and Da Ma contribute equally to this work.} \quad
  Da Ma$^1\:\!^*$ \quad
  Lu Chen$^{1,\:\!2}$\:\!\thanks{Xin Chen, Kai Yu, and Lu Chen are the corresponding authors.} \quad
  Liangtai Sun$^1$ \quad
  Zihao Li$^3$ \quad
  Yi Xia$^2$ \\
  \textbf{Bo Chen}$^2$ \quad
  \textbf{Hongshen Xu}$^1$ \quad
  \textbf{Zichen Zhu}$^1$ \quad
  \textbf{Su Zhu}$^4$ \quad
  \textbf{Shuai Fan}$^4$ \quad
  \textbf{Guodong Shen}$^2$ \\
  \textbf{Kai Yu}$^{1,\:\!2\:\!\dag}$ \quad
  \textbf{Xin Chen}$^2\:\!^\dag$ \\
  $^1$X-LANCE Lab, Department of Computer Science and Engineering\\
  MoE Key Lab of Artificial Intelligence, SJTU AI Institute\\
  Shanghai Jiao Tong University, Shanghai, China \\
  $^2$Suzhou Laboratory, Suzhou, China\\
  $^3$Shanghai Key Laboratory for Molecular Engineering of Chiral Drugs\\
  School of Chemistry and Chemical Engineering\\
  Shanghai Jiao Tong University, Shanghai, China\\
  $^4$AI Speech Co, .Ltd., Suzhou, China\\
  \texttt{\{zhao\_mengxin, chenlusz, kai.yu\}@sjtu.edu.cn} \\
}

\begin{document}

\maketitle

\begin{abstract}

Artificial intelligence (AI) has played an increasingly important role in chemical research. However, most models currently used in chemistry are specialist models that require training and tuning for specific tasks. A more generic and efficient solution would be an AI model that could address many tasks and support free-form dialogue in the broad field of chemistry. In its utmost form, such a generalist AI chemist could be referred to as Chemical General Intelligence. Large language models (LLMs) have recently logged tremendous success in the general domain of natural language processing, showing emerging task generalization and free-form dialogue capabilities. However, domain knowledge of chemistry is largely missing when training general-domain LLMs. The lack of such knowledge greatly hinders the performance of generalist LLMs in the field of chemistry. To this end, we develop ChemDFM, a pioneering LLM for chemistry trained on 34B tokens from chemical literature and textbooks, and fine-tuned using 2.7M instructions. As a result, it can understand and reason with chemical knowledge in free-form dialogue. Quantitative evaluations show that ChemDFM significantly surpasses most representative open-source LLMs. It outperforms GPT-4 on a great portion of chemical tasks, despite the substantial size difference. We have open-sourced the inference codes, evaluation datasets, and model weights of ChemDFM on Huggingface\footnote{\url{https://huggingface.co/OpenDFM/ChemDFM-v1.0-13B}}.

\end{abstract}

\section{Introduction}

With the rapid development of artificial intelligence (AI), utilizing AI to assist chemical research has garnered increasing attention~\citep{wang2023scientific,back2024accelerated}. 
Various AI models have been developed for tasks such as property prediction~\citep{zhou2022uni,wu2023molformer,chen2023geomformer},
molecular captioning and generation~\citep{xu2021geodiff,edwards-etal-2022-translation,perron2022deep,du2024machine,lu2024d}, and reaction predictions~\citep{schwaller2020predicting,wang2021retroprime,han2024retrosynthesis}. 
Since BERT~\citep{devlin-etal-2019-bert} and GPT~\citep{radfordimproving}, efforts have been made to fine-tune pre-trained models for specific chemical tasks~\citep{zhou2022uni,edwards-etal-2022-translation,liu-etal-2023-molxpt,luo2023one,zhang2024unimot}.
However, these models are typically trained on a meticulously curated dataset to solve a designated task in a particular scenario, leading to a one-to-one relationship between models and tasks. Once out of that specific scenario, they are often not useful, even for highly related tasks. 
A more attractive and practical AI system should be capable of handling a wide range of chemical tasks under real-world scenarios and conducting free-form human-AI collaborations.
Such an AI system necessitates a comprehensive array of chemical competencies, coupled with the ability to comprehend and reason in both chemical and natural languages. This would enable it
to work as a research assistant or even collaborator alongside human researchers. This could be an essential step towards eventually achieving Chemical Artificial General Intelligence. 

In pursuit of a highly integrated AI system for a broad range of chemical challenges, recent advancements in large language models (LLMs)~\citep{du-etal-2022-glm,touvron2023llama,xu-etal-2023-baize} brought great new hopes. Numerous studies have demonstrated the remarkable competencies of LLMs in natural language understanding and task generalization~\citep{wei2021finetuned,xu-etal-2023-baize}, deductive reasoning~\citep{wei2022chain,kojima2022large}, and tool utilization~\citep{schick2023toolformer,qin2024toolllm}. These made LLMs shine in traditional natural language processing tasks and accomplish problems that were previously unimaginable and unsolvable, such as handling tasks in unknown scenarios or conducting free-form dialogues with humans. These inherent strengths underscore the viability of employing LLMs as AI-driven research collaborators in the field of chemistry.

\begin{wrapfigure}[]{r}{0.5\linewidth}
    \centering
    \includegraphics[width=1.0\linewidth]{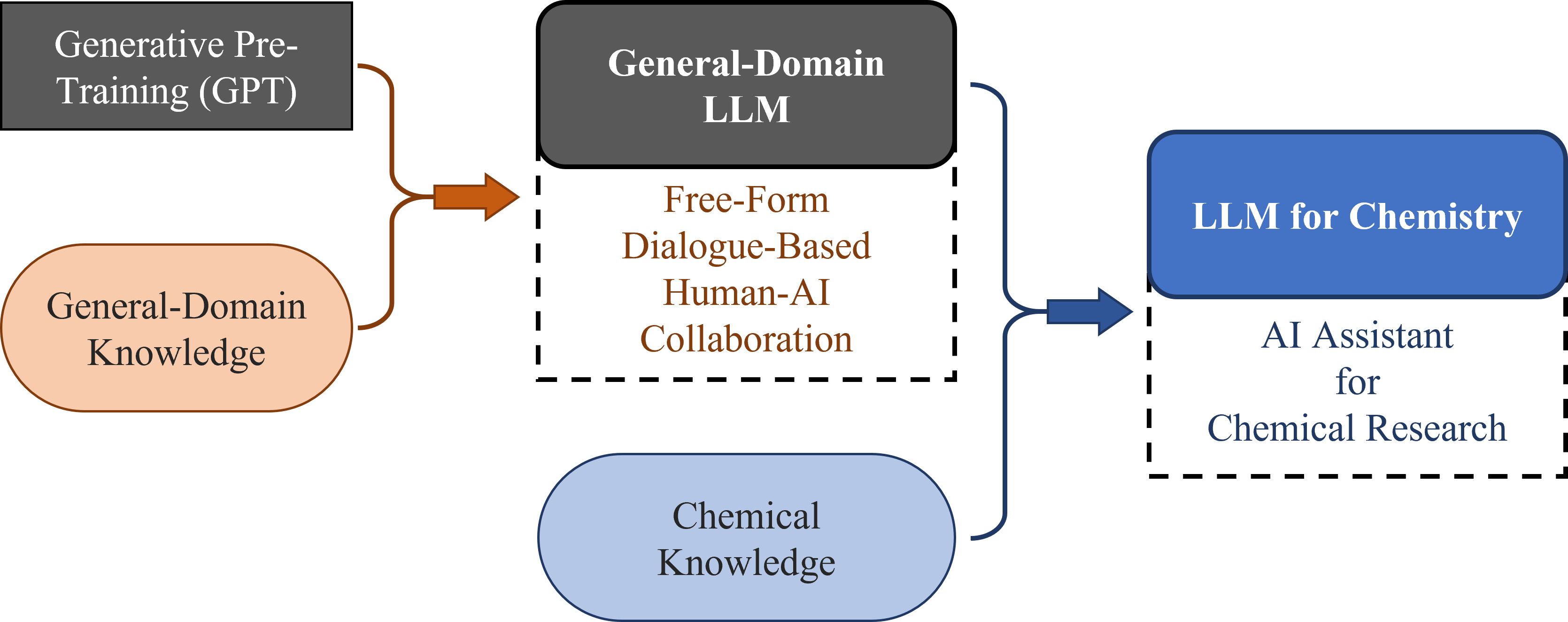}
    \caption{Scheme to obtain an LLM for chemistry, through using chemical domain knowledge to train a general-domain LLM. }\label{fig:relation}
\end{wrapfigure}

Different from general domains, tasks in chemical domains necessitate models to possess additional chemical comprehension capabilities for understanding and reasoning over chemical-specialized language and knowledge.
This hinders general domain LLMs from excelling in chemical tasks as they often lack in-depth chemical knowledge~\citep{kristiadi2024sober}. For example, molecules are a vital component of the chemical world. Although molecules can be conveyed through natural-language-like notations such as SMILES (Simplified Molecular Input Line Entry System), IUPAC names, and molecular formulas, their meanings and intrinsic structures are entirely different from those in natural language. CO represents carbon monoxide in chemistry, not Colorado, while Co represents Cobalt, not a company, and (CO) as part of a SMILES typically represents the carbonyl group. The lack of understanding of these molecular notations severely limits the applicability and performance of general domain LLMs in solving chemistry problems. Therefore, we believe that equipping general-domain LLMs with rich chemical knowledge of task-specific chemical models, as illustrated in Figure~\ref{fig:relation}, is vital for developing LLMs useful in the field of chemistry.


In this work, we propose \textbf{ChemDFM}, a Dialogue Foundation Model for Chemistry.
ChemDFM takes advantage of the pre-trained LLaMa-13B model~\citep{touvron2023llama}, an open-source general-domain LLM, and is further specialized in chemistry through two phases: 1) \textbf{Domain Pre-training}, where the model harvests the chemical knowledge from research articles and textbooks, and 2) \textbf{Instruction Tuning}, where the model familiarizes itself with chemical language and patterns, especially molecule notations. Each phase uses an extensive and diverse collection of chemical data:
1) nearly 34B tokens from over 3.8M chemical papers and 1.4K textbooks in chemistry used in Phase I, and 2) over 2.7M instructions crafted from various chemical databases in Phase II.
Apart from chemical data, we also incorporated a substantial amount of general-domain data in both phases to make sure that ChemDFM maintains comprehension and reasoning capabilities of natural language while acquiring new chemical knowledge. As a result, ChemDFM can simultaneously handle a wide range of chemical tasks and convey free-form dialogues using the language of chemists, enabling human-AI collaboration in chemical research.

A series of experiments have been conducted to evaluate the prowess of ChemDFM, including molecule recognition, molecule design, molecular property prediction, and reaction analysis.
The results show that ChemDFM achieves advanced performances, surpassing typical open-source LLMs. It even outperforms GPT-4 on most tasks, despite the notable difference in model size.
We further compared the performance between ChemDFM and the baseline LLMs in free-form unseen scenarios analogues to real-world scenarios. The test examples were constructed based on the latest chemical papers to avoid possible data leakage. The results show that ChemDFM can generate answers that are more accurate and relevant to the specific questions. These findings suggest
that ChemDFM, capable of handling a broad range of chemical tasks and reasoning in both chemical and natural languages, can indeed serve as an AI assistant in chemical research.

\section{ChemDFM}

\begin{figure*}
    \centering
    \includegraphics[width=\linewidth]{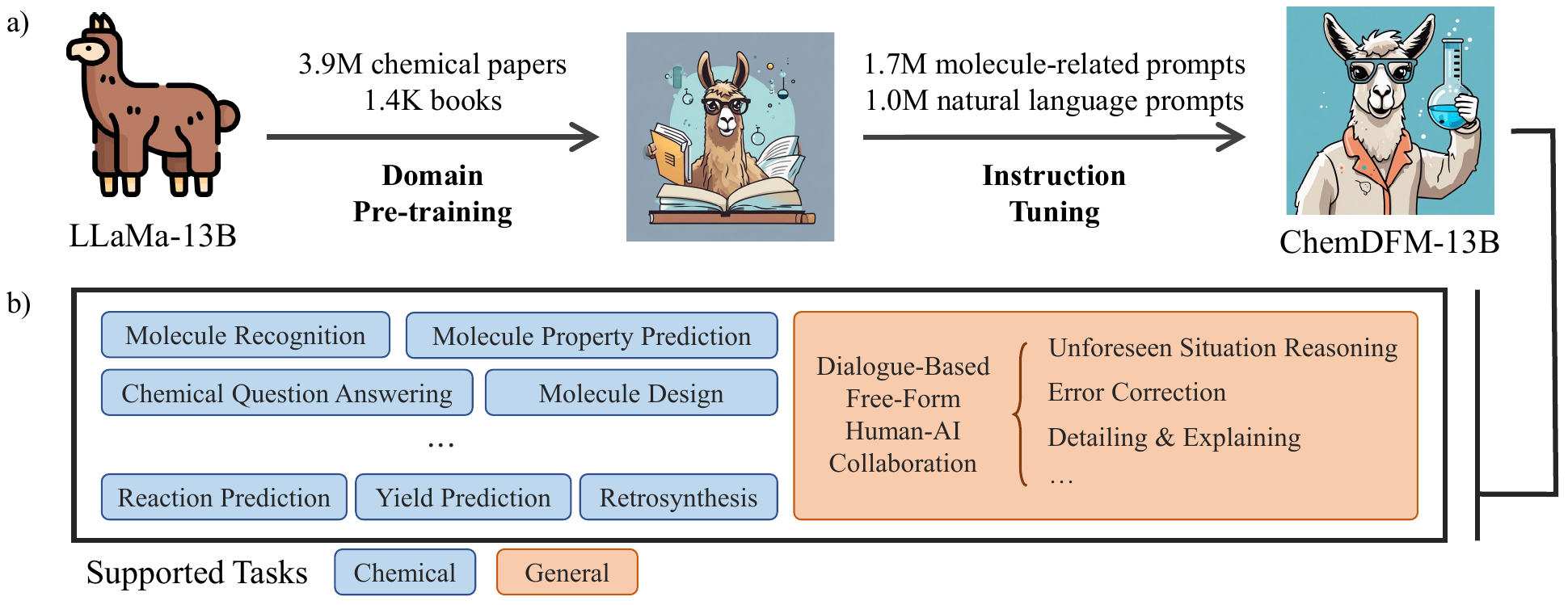}
    \caption{a) Two-step training procedure to obtain ChemDFM. The icons are generated by the SDXL model provided by Stability AI\protect\footnotemark. b) Various types of tasks ChemDFM is capable of. }
    \label{fig:main}
    \vspace{-3mm}
\end{figure*}

As outlined in Figure~\ref{fig:main}, ChemDFM is trained based on LLaMa, a general domain LLM. Domain knowledge of chemistry is instilled in ChemDFM in two steps: Domain Pre-training and Instruction Tuning. Through this two-stage specialization process, ChemDFM “learned” chemistry and gained abilities such as molecule recognition and reaction prediction. The training process is presented below and evaluations of ChemDFM’s capability are elaborated in the next section.

\footnotetext{\url{https://stability.ai/}}

\subsection{Domain Pre-training}\label{sec:pretrain}

Data used to train general-domain LLMs must contain knowledge covering a wide range of topics. Such broadness is often accompanied by sacrifices of deepness in each field.
While models trained on such data have successfully gained strong natural language understanding and reasoning capabilities, they often fall short when it comes to in-depth specialized knowledge. The lack of domain knowledge is partially responsible for the well-known “hallucination” problem~\citep{huang2023surveyhallucinationlargelanguage}. To alleviate this problem, we collected a corpus of data rich in chemical knowledge for domain pre-training, primarily from the two most authoritative sources for chemical knowledge: textbooks and published papers.
Textbooks represent the widely accepted knowledge and basic principles of chemistry while published papers offer more details and more up-to-date chemical knowledge, some of which have not been incorporated into textbooks.
Specifically, we selected 1.4K chemistry books from LibreTexts\footnote{\url{https://libretexts.org/}} and Gold Books\footnote{\url{https://goldbook.iupac.org/}} and collected 3.9M open-access papers in chemistry-related topics before January 2022. After further pre-processing and deduplication, we obtained 49M tokens from the textbooks and 34B tokens from the published research articles.
To maintain the LLM's general-domain knowledge and capabilities, we also included highly selective data in the general field, including Wikipedia, Arxiv, Books, StackExchange, GitHub code, WuDao Corpora~\citep{yuan2021wudaocorpora}, etc.
More details of domain pre-training are available in Appendix Section~\ref{sec:dp}.

\subsection{Instruction Tuning}\label{sec:instune}

The data for the chemical instruction tuning dataset comprises two main categories: chemical knowledge presented in natural language and specialized molecular notations.
%
A dataset containing over 1M question-answering pairs specialized in chemistry was constructed for instruction turning to enhance the model’s capability to process chemistry-related natural language.
These data were generated from two sources. The first one is established question-answering datasets, including {\tt ARC}~\citep{clark2018think}, {\tt PIQA}~\citep{bisk2020piqa}, {\tt SciQ}~\citep{welbl2017crowdsourcing}, and {\tt HendrycksTest}~\citep{hendryckstest2021}. The other source of questions is middle school exams. We collected open-source exam questions from the Internet and constructed question-answer pairs (with key points or problem-solving strategies when available) for the instruction tuning of ChemDFM.

While natural languages such as English or German are generally descriptive and highly versatile, they are often not the best media to convey chemical knowledge. For example, it is often much easier and more comprehensible to draw the molecular structure of a complicated organic molecule than to describe it using natural language. Generations of chemists have developed many specialized notations, such as molecular formulas and Simplified Molecular Input Line Entry System~(SMILES)~\citep{weininger1988smiles} notation. This represents a key challenge for LLMs to understand chemistry. A key goal of the instruction tuning stage was to tackle this challenge by familiarizing ChemDFM with the specialized notations. In training ChemDFM, we chose SMILES, one of the most popular notations of molecules, as the main representation for molecules. It uses a sequence of letters to present a molecule, retaining rich structural information such as molecular configuration and chirality in most cases. In addition, its text-like data structure makes it highly compatible with LLMs.

\begin{wraptable}[29]{r}{0.6\linewidth}
    \centering
    \vspace{-3mm}
    \caption{Itemized list of our instruction tuning dataset. MD: Molecule Description, TBMD: Text-Based Molecule Design, MPP: Molecular Property Prediction, RC: Reaction Completion, MNA: Molecular Notation Alignment.}
    \label{tab:data}
    \begin{tabular}{ccc}
    \toprule
        Data Type & \# samples & Data Source \\
    \midrule
        \multirow{2}{*}{QAs from Datasets} & \multirow{2}{*}{131K} & ARC, PIQA, SciQ \\
         & & HendrycksTest \\
        QAs from Exam & 915K & Internet \\
        MD & 576K & PubChem \\
        TBMD & 576K & PubChem \\
        MPP & 102K & MoleculeNet \\
        RC & 300K & USPTO \\
        MNA & 120K & PubChem \\
    \bottomrule
    \end{tabular}
\end{wraptable}

To help the model comprehend SMILES, three kinds of molecular data were used:
1. \textbf{Molecule description~(MD) and text-based molecule design~(TBMD).} Our dataset includes all the SMILES-description pairs from PubChem\footnote{\url{https://pubchem.ncbi.nlm.nih.gov/}}, a web-scale chemical database that contains more than 100M compounds.
The model was instructed to generate descriptions of given molecules or reversely, generate molecule(s) that match a description. We duplicated samples with descriptions longer than two sentences to further enhance the data quality.
2. \textbf{Molecular property prediction~(MPP)}. The model was instructed to predict the properties of a given molecule. This data was constructed based on the widely used molecular property prediction benchmark, MoleculeNet~\citep{wu2018moleculenet}.
3. \textbf{Reaction completion~(RC)}.
The model was also instructed to complete chemical reactions in which one or more reactants/products were masked randomly. The reactions were sampled from USPTO~\citep{lowe2012extraction}, the largest open-source chemical reaction database.

\begin{wrapfigure}[1]{r}{0.6\linewidth}
    \centering
    \vspace{-55mm}
    \includegraphics[width=\linewidth]{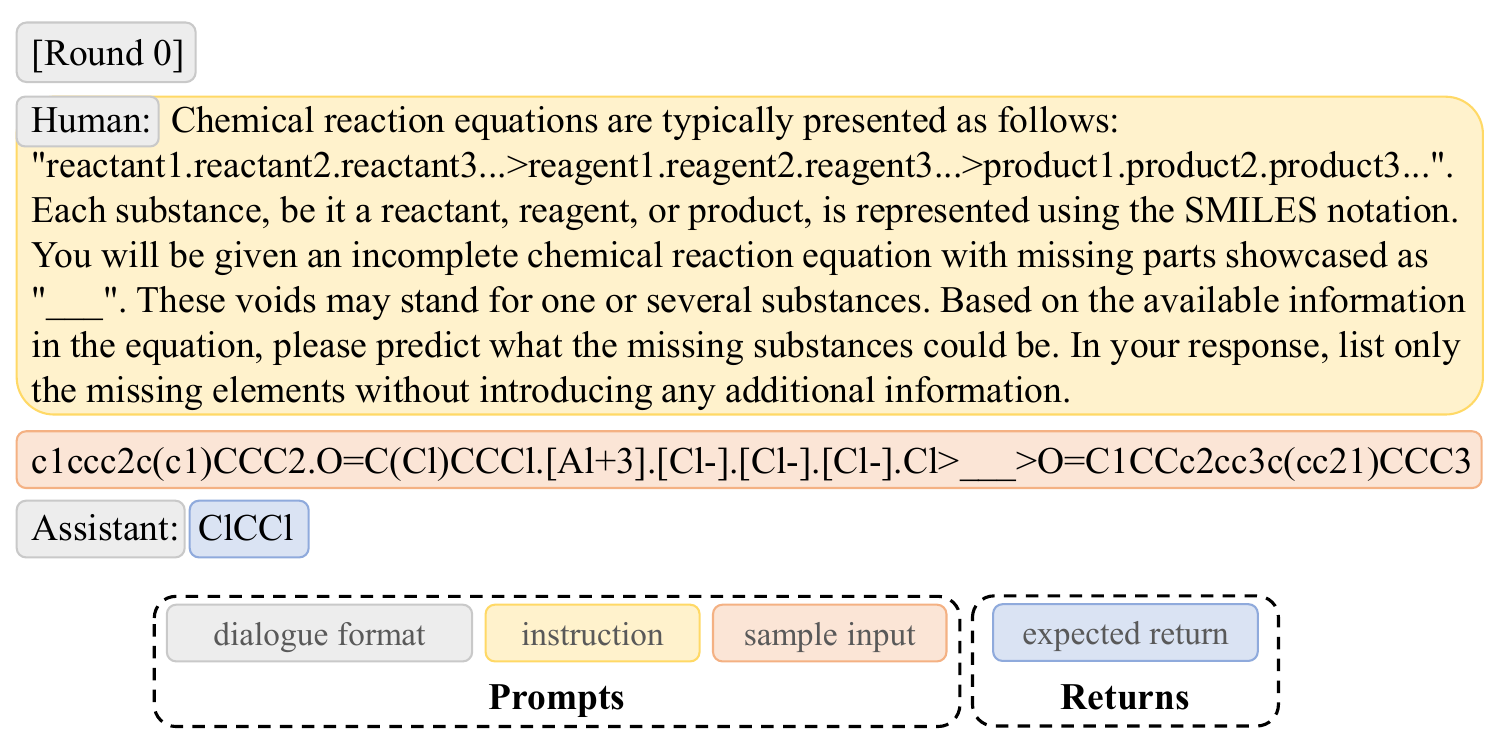}
    \caption{Representative question used in instruction tuning.}
    \label{fig:data}
\end{wrapfigure}

In addition to SMILES, we indirectly include two other widely used notations of molecules, IUPAC names and molecular formulas. We instructed the model to translate between the three notations, e.g. predicting SMILES of a molecule given its IUPAC name and vice versa, allowing it to understand these alternatives. This kind of data is called \textbf{Molecular Notation Alignment~(MNA)} in this work.

Table~\ref{tab:data} lists the itemized entries of our instruction tuning dataset. All the data samples take the form of $(\mathtt{prompt}, \mathtt{returns})$ tuples, where the $\mathtt{prompt}$ is composed of the dialogue format, instructions, and example inputs, and the $\mathtt{returns}$ are the expected outputs. Such an example is presented in Figure~\ref{fig:data}. To diversify the natural language instructions, we used GPT-4 to rephrase instructions for all tasks. The number of different instructions for each task ranges from 20 to 200.

To maintain the advanced natural language comprehension abilities, we also included a substantial amount of general domain data for the instruction-tuning of ChemDFM. The ratio of data from the chemical domain to the general domain is roughly 1:2.
The instruction tuning of ChemDFM is a full-parameter tuning process with more details in Appendix Section~\ref{sec:it}.

\section{Evaluations}\label{sec:obj}

To assess ChemDFM’s capability in chemistry, we compared its performance against several generalist LLM models: GPT-4~\citep{openai2023gpt4}, LLaMa-2~\citep{touvron2023llama2} and Galactica~\citep{taylor2022galactica}, as they represent very large generalist LLMs, medium-sized generalist LLMs and LLMs tuned for science, respectively. We used ChemLLMBench~\citep{guo2023large} for quantitative evaluation of ChemDFM’s ability in chemistry and then carried out qualitative analyses of ChemDFM’s free-form collaboration capacity, focusing on its superior chemistry-related conversation processing power.

\subsection{Quantitative Evaluation}\label{benchmarks}

ChemLLMBench is made of a series of chemical tasks covering a wide range of chemistry-related topics. The standard form of evaluation was conducted on 100 instances randomly sampled from the respective test sets of the tasks. To ensure a fair comparison, we used the same 100 samples when comparing different LLMs, unless otherwise specified. Some non-LLM task-specific models were used for comparisons when available. Detailed explanations of the tasks and the prompt format for ChemDFM can be found in Appendix Section~\ref{sec:quantitative}. Specifically, the quantitative evaluation tasks can be categorized into the following four groups.

\begin{wraptable}[16]{r}{0.6\linewidth}
    \centering
    \vspace{-5mm}
    \caption{Accuracy scores in name prediction tasks. Baseline results are from \citet{guo2023large}. S2I: SMILES to IUPAC names translation, I2S: IUPAC names to SMILES translation, S2MF: SMILES to molecule formulas translation, I2MF: IUPAC names to molecule formulas translation.}
    \label{tab:moltrans}
    \begin{tabular}{lcccc}
    \toprule
    Model & S2I$\uparrow$ & I2S$\uparrow$ & S2MF$\uparrow$ & I2MF$\uparrow$  \\
    \midrule
    \rowcolor{grey}\multicolumn{5}{c}{\textit{Task-specific specialist models}} \\
    STOUT & 55.0 & 70.0 & - & - \\
    \midrule
    \rowcolor{grey}\multicolumn{5}{c}{\textit{LLM-based generalist models}} \\
    GPT-4 & 0 & 1.2 & 8.6 & 8.4 \\
    LLaMa2-13B-chat & 0 & 0 & 1.0 & 0 \\
    Galactica-30B & 0 & 0 & 0 & 0 \\
    \textbf{ChemDFM-13B} & \textbf{4.0} & \textbf{11.0} & \textbf{73.0} & \textbf{51.0} \\
    \bottomrule
    \end{tabular}
\end{wraptable}

\paragraph{1) Molecule recognition.}
There are two series of tasks in ChemLLMBench that directly assess the capability to recognize molecules: \textit{name prediction} and \textit{molecule captioning}.
In the \textit{name prediction} tasks, a model is asked to translate between different notations for molecules, including SMILES, IUPAC name, and molecular formula. Specifically, it consists of four tasks: SMILES to IUPAC name translation~(S2I), IUPAC name to SMILES translation~(I2S), SMILES to Molecular Formula translation~(S2MF), and IUPAC name to Molecular Formula translation~(I2MF). For IUPAC names and SMILES, we normalized the predictions before calculating the accuracy scores, while for molecular formulas, only exact matches are considered correct answers. The \textit{molecule captioning} tasks further require LLMs to not only recognize the molecule present by a given SMILES notation but also generate a brief description of it using natural language. In these tasks, traditional captioning metrics like BLUE, ROUGE, and METEOR are used to assess the model’s performance on a test set of ChEBI-20~\citep{edwards-etal-2021-text2mol}.

Benchmark performance of different models on these two molecule recognition tasks is reported in Table~\ref{tab:moltrans} and Table~\ref{tab:description}, respectively. Table 2 shows that most LLMs, including GPT-4, can hardly complete \textit{name prediction} tasks, indicating a limited understanding of molecules and ChemDFM outperforms open-source LLMs by a significant margin across all these tasks. This outstanding performance of ChemDFM proves its robust molecule recognition capabilities and validates the effectiveness of our specialization process.

In \textit{molecule captioning} tasks (as shown in Table~\ref{tab:description}), ChemDFM also performs far superior to open-source LLMs. The results denote that ChemDFM not only recognizes molecules but also infers their underlying chemical essence and nature.
It is worth noting the drastic drop in GPT-4’s performance from the ten-shot setting to the zero-shot setting, which indicates that GPT-4 thrives mostly on its extraordinary natural language capabilities to learn from given exemplars while its inherent molecule recognition capability is relatively fragile. Comparatively, ChemDFM achieves comparable performance without the help of exemplars, demonstrating its intrinsic molecule recognition capability.

\begin{table*}[t]
    \centering
    \caption{Benchmark results of different models in molecule captioning tasks. \dag: results from \citet{guo2023large}. *: reproduced results.}
    \label{tab:description}
    \resizebox{\textwidth}{!}{
    \begin{tabular}{lcccccc}
    \toprule
    Model & BLEU-2$\uparrow$ & BLEU-4$\uparrow$ & ROUGE-1$\uparrow$ & ROUGE-2$\uparrow$ & ROUGE-L$\uparrow$ & METEOR$\uparrow$ \\
    \midrule
    \rowcolor{grey}\multicolumn{7}{c}{\textit{Task-specific specialist models}} \\
    Text+Chem T5~\citep{christofidellis2023unifying} & 0.625 & 0.542 & 0.682 & 0.543 & 0.622 & 0.648 \\
    MolXPT~\citep{liu-etal-2023-molxpt} & 0.594 & 0.505 & 0.660 & 0.511 & 0.597 & 0.626 \\
    InstructMol~\citep{cao2023instructmol} & 0.475 & 0.371 & 0.566 & 0.394 & 0.502 & 0.509 \\
    Mol-Instruction~\citep{fang2023molinstructions} & 0.249 & 0.171 & 0.331 & 0.203 & 0.289 & 0.271 \\
    \midrule
    \rowcolor{grey}\multicolumn{7}{c}{\textit{LLM-based generalist models}} \\
    GPT-4~(10-shot)\textsuperscript{\dag} & \textbf{0.464} & \textbf{0.365} & \textbf{0.545} & \underline{0.362} & \underline{0.459} & \textbf{0.519} \\
    GPT-4~(0-shot)\textsuperscript{\dag} & 0.062 & 0.013 & 0.192 & 0.040 & 0.125 & 0.209 \\
    LLaMa-2-13B-chat~(10-shot)\textsuperscript{\dag} & 0.197 & 0.140 & 0.331 & 0.193 & 0.265 & 0.372 \\
    Galactica-30B~(10-shot)\textsuperscript{*} & 0.114 & 0.055 & 0.334 & 0.189 & 0.330 & 0.187 \\
    Galactica-30B~(0-shot)\textsuperscript{\dag} & 0.008 & 0.002 & 0.019 & 0.004 & 0.015 & 0.043 \\
    \textbf{ChemDFM-13B~(0-shot)} & \underline{0.321} & \underline{0.265} & \underline{0.490} & \textbf{0.374} & \textbf{0.483} & \underline{0.402} \\
    \bottomrule
    \end{tabular}
    }
    \vspace{-3mm}
\end{table*}

\paragraph{2) Text-based molecule design.} To evaluate LLM’s efficiency in making qualified molecule designs, ChemLLMBench reverses the molecule captioning tasks and asks the models to generate molecules based on their descriptions. Specifically, in the \textit{text-based molecule design} task, models are asked to predict the SMILES of the molecule that fits the given description. Two sets of metrics are utilized to measure the performance of these tasks. The first set measures the text-based similarity of the predicted SMILES compared to the golden SMILES, which includes exact match, BLUE, and Levenshtein distance. The second set of metrics measures the chemical similarity of the predicted molecules to the golden molecules, including the validity of the predicted SMILES and the FTS~(fingerprint Tanimoto Similarity)~\citep{tanimoto1958elementary} in terms of MACCS~\citep{maccs}, RDK\footnote{\url{https://www.rdkit.org/}}, Morgan~\citep{morgan}.

\begin{table*}[t]
    \centering
    \caption{Benchmark results of different models in text-based molecule design tasks. \dag: results from \citet{guo2023large}. *: 10-shot results}
    \label{tab:design}
    \resizebox{\textwidth}{!}{\begin{tabular}{lccccccc}
    \toprule
    Model & Exact$\uparrow$ & BLEU$\uparrow$ & Dis$\downarrow$ & Validity$\uparrow$ & MACCS$\uparrow$ & RDK$\uparrow$ & Morgan$\uparrow$ \\
    \midrule
    \rowcolor{grey}\multicolumn{8}{c}{\textit{Task-specific specialist models}} \\
    MolXPT~\citep{liu-etal-2023-molxpt} & 21.5 & - & - & 98.3 & 0.859 & 0.757 & 0.667 \\
    Text+Chem T5~\citep{christofidellis2023unifying} & 32.2 & 0.853 & 16.87 & 94.3 & 0.901 & 0.816 & 0.757 \\
    Mol-Instruction~\citep{fang2023molinstructions} & 0.2 & 0.345 & 41.4 & 100 & 0.412 & 0.231 & 0.147 \\
    \midrule
    \rowcolor{grey}\multicolumn{8}{c}{\textit{LLM-based generalist models}} \\
    GPT-4\textsuperscript{\dag}\textsuperscript{*} & 17.4 & 0.816 & 21.2 & 88.8 & 0.867 & 0.738 & 0.672 \\
    LLaMa-2-13B-chat\textsuperscript{\dag}\textsuperscript{*} & 2.0 & 0.626 & 34.0 & 78.2 & 0.679 & 0.568 & 0.454 \\
    Galactica-30B\textsuperscript{\dag} & 0.0 & 0.004 & 2738 & 95.6 & 0.233 & 0.109 & 0.053 \\
    \textbf{ChemDFM-13B} & \textbf{45.0} & \textbf{0.874} & \textbf{9.9} & \textbf{98.0} & \textbf{0.922} & \textbf{0.871} & \textbf{0.798} \\
    \bottomrule
    \end{tabular}}
    \vspace{-3mm}
\end{table*}

As shown in Table~\ref{tab:design}, ChemDFM outperforms not only the generalist LLMs but also the traditional task-specific specialist models across almost all metrics, which is both surprising and promising. Considering that task-specific specialist models were evaluated on the entire test set, whereas the performance of ChemDFM was initially assessed on only 100 samples, we further evaluated ChemDFM on the complete test set to align with the task-specific models for a fair comparison. The results, shown in Table~\ref{tab:full} of the Appendix, further validate the advantage of ChemDFM. The results from Table~\ref{tab:design} and \ref{tab:full} unveil two key superiorities of ChemDFM over other models.
On the one hand, ChemDFM has effectively established a relationship between SMILES notations and the chemical nature of compounds in our model, which other LLMs lack. On the other hand, ChemDFM benefits from the solid natural language comprehension capabilities inherited from LLaMa, which task-specific specialist models lack. Altogether, ChemDFM constructs a more comprehensive knowledge system in chemistry, which helps it surpass both generalist and task-specific specialist models.

\paragraph{3) Molecular property prediction.}
The \textit{molecular property prediction} tasks in ChemLLMBench consist of five tasks from the MoleculeNet~\citep{wu2018moleculenet}, including BACE, BBBP, HIV, ClinTox, and Tox21. Among them, BACE and BBBP each contain a single balanced binary classification task. HIV contains a single unbalanced binary classification task. ClinTox and Tox21 comprise two and twenty-one unbalanced binary classification tasks, respectively.
To address the severe label imbalance in these tasks, the Area Under the Curve of the Receiver Operating Characteristic~(AUC-ROC) metric~\citep{bradley1997use} was introduced. To better assess the molecular property prediction, we adopted a scaffold-vertical manner for data splitting. Specifically, the molecules from the DeepChem library~\citep{deepchem} were first grouped based on their Bemis-Murcko scaffold~\citep{bemis1996properties} representations. 
The datasets were then split into training and test sets according to these groups. This method ensures that no molecule sharing the same scaffold would appear in both the training set and the test set. While avoiding information leaking due to mere similarity of molecules, this method also significantly increases the difficulty of the tasks, making the assessment more challenging and meaningful.
The results listed in Table~\ref{tab:molnet} show that ChemDFM consistently outperforms other LLMs in all but one molecular property prediction task. 

\begin{table}[t]
    \centering
    \caption{AUC-ROC scores~\citep{bradley1997use} of different models in molecular property prediction tasks. Avg: average. \dag: reproduced results (The results of GPT-4 were obtained in January 2024).}
    \label{tab:molnet}
    \begin{tabular}{lcccccc}
    \toprule
    Model & BACE$\uparrow$ & BBBP$\uparrow$ & ClinTox$\uparrow$ & HIV$\uparrow$ & Tox21$\uparrow$ & Avg$\uparrow$ \\
    \midrule
    \rowcolor{grey}\multicolumn{7}{c}{\textit{Task-specific specialist models}} \\
    Uni-Mol~\citep{zhou2022uni} & 85.7 & 72.9 & 91.9 & 80.8 & 79.6 & 82.2 \\
    MolXPT~\citep{liu-etal-2023-molxpt} & 88.4 & 80.0 & 95.3 & 78.1 & 77.1 & 83.8 \\
    InstructMol~\citep{cao2023instructmol} & 85.9 & 64.0 & - & 74.0 & - & - \\
    \midrule
    \rowcolor{grey}\multicolumn{7}{c}{\textit{LLM-based generalist models}} \\
    GPT-4\textsuperscript{\dag} & 62.5 & 61.5 & 51.6 & 65.9 & 55.2 & 59.3 \\
    LLaMa-2-13B-chat\textsuperscript{\dag} & 26.0 & 60.3 & 45.7 & 29.0 & 51.7 & 42.5 \\
    Galactica-30B~\citep{taylor2022galactica} & 72.7 & 59.6 & 82.2 & \textbf{75.9} & 68.5 & 71.8 \\
    \textbf{ChemDFM-13B} & \textbf{78.4} & \textbf{66.7} & \textbf{89.9} & 73.6 & \textbf{79.8} & \textbf{77.7} \\
    \bottomrule
    \end{tabular}
    \vspace{-3mm}
\end{table}

\begin{table}[t]
    \centering
    \caption{Accuracy scores of different models in reaction prediction and retrosynthesis tasks. B-H: Buchwald-Hartwig dataset~\citep{ahneman2018predicting}. Suzuki: Suzuki-Miyaura dataset~\citep{reizman2016suzuki}. YP: Yield Prediction, RP: Reactant Prediction, RS: Reagent Selection, Retro: Retrosynthesis. \dag: results from \citet{guo2023large}. Please refer to Table \ref{tab:yp}$\sim$\ref{tab:rs} in the Appendix for complete results.}
    \label{tab:reaction}
    \begin{tabular}{lcccc}
    \toprule
    Model & YP$\uparrow$ & RP$\uparrow$ & Retro$\uparrow$ & RS$\uparrow$ \\
    \midrule
    \rowcolor{grey}\multicolumn{5}{c}{\textit{task-specific specialist models}} \\
    UAGNN~\citep{kwon2022uncertainty} & 96.1 & - & - & - \\
    Chemformer~\citep{irwin2022chemformer} & - & 93.8 & 53.6 & - \\
    \midrule
    \rowcolor{grey}\multicolumn{5}{c}{\textit{LLM-based generalist models}} \\
    GPT-4\textsuperscript{\dag} & \underline{78.2} & \underline{23.0} & \underline{11.4} & \textbf{45.3} \\
    LLaMa-2-13B-chat\textsuperscript{\dag} & 0.7 & 3.2 & 0.0 & 16.0 \\
    Galactica~(30B)\textsuperscript{\dag} & 0.4 & 3.6 & 1.6 & 8.0 \\
    \textbf{ChemDFM-13B} & \textbf{81.0} & \textbf{49.0} & \textbf{12.0} & \underline{23.7} \\
    \bottomrule
    \end{tabular}
    \vspace{-3mm}
\end{table}

\paragraph{4) Reaction prediction and retrosynthesis.}
ChemLLMBench includes four types of tasks targeted at evaluating models' capability of reaction understanding: \textit{Yield Prediction}~(YP), \textit{Reaction Prediction}~(RP), \textit{Reagent Selection}~(RS), and \textit{Retrosynthesis}~(Retro).
The \textit{yield prediction} tasks ask models to predict whether the given reaction is a high-yield reaction and are constructed based on two High-Throughput experimentation (HTE) datasets: the Buchwald-Hartwig dataset~\citep{ahneman2018predicting} and the Suzuki-Miyaura dataset~\citep{reizman2016suzuki}. The \textit{reaction prediction} tasks ask models to predict the product of the given reaction. The USPTO-MIT dataset~\citep{jin2017predicting} was used to construct these tasks. The \textit{retrosynthesis} tasks focus on predicting the reactants of the given reactions and are constructed based on the USPTO-50K dataset~\citep{schneider2016s}. The \textit{reagent selection} tasks focus on selecting the appropriate reactants, solvents, or ligands that lead to a higher yield of the reaction from a list of candidates
based on the dataset proposed by \citet{perera2018platform}. Accuracy is utilized to measure the performance.

As depicted in Table~\ref{tab:reaction}, ChemDFM’s performance significantly exceeds open-source LLMs in all the reaction-related tasks.
When compared to GPT-4, ChemDFM achieved superior performances on 3 out of 4 tasks, but lagged behind on reagent selection tasks. A closer inspection reveals that these reagent selection tasks are multiple-choice questions where models are asked to directly copy the correct SMILES from the candidates listed in the questions. Compared with generative questions such as reaction prediction and retrosynthesis, multiple-choice questions alleviate the models' burden of generating molecules from scratch. It seems that GPT-4 can indeed better follow the instructions and directly copy corresponding SMILES, while ChemDFM often tries to generate new answers.

\begin{figure*}[t]
    \centering
    \includegraphics[width=\textwidth]{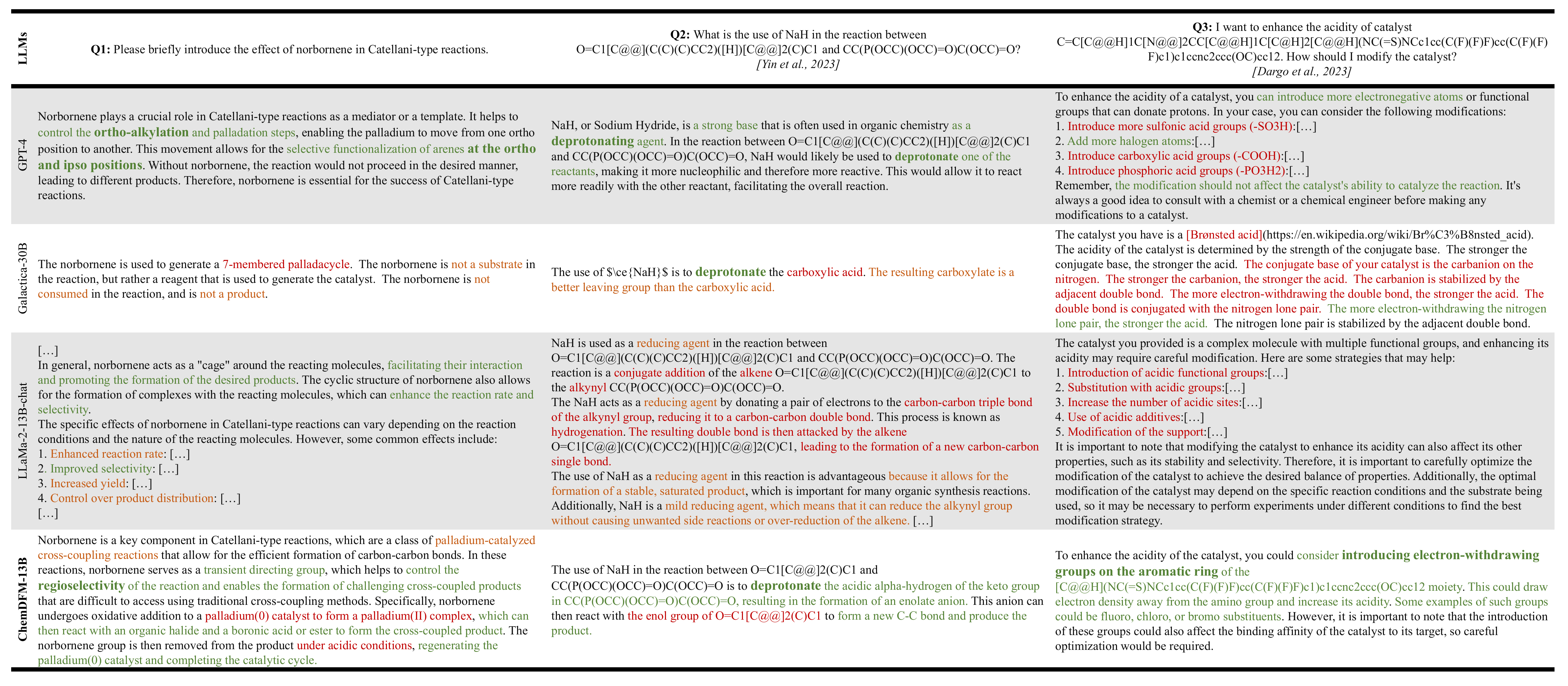}
    \caption{Examples of paper reading. Answers from ChemDFM are compared with GPT and the base model LLaMa. \textcolor{green}{Correct and relevant information} in the replies is marked in green, \textcolor{yellow}{correct but irrelevant information} in yellow, and \textcolor{red}{wrong information} in red. \textbf{Key points of the answer} are marked in bold. Full details and more examples are elaborated in Appendix Section~\ref{sec:pr}.}
    \label{fig:one}
    \vspace{-5mm}
\end{figure*} 

\subsection{LLM-Based Research Assistant}

\begin{wrapfigure}[21]{r}{0.65\linewidth}
    \centering
    \vspace{-5mm}
    \includegraphics[width=\linewidth]{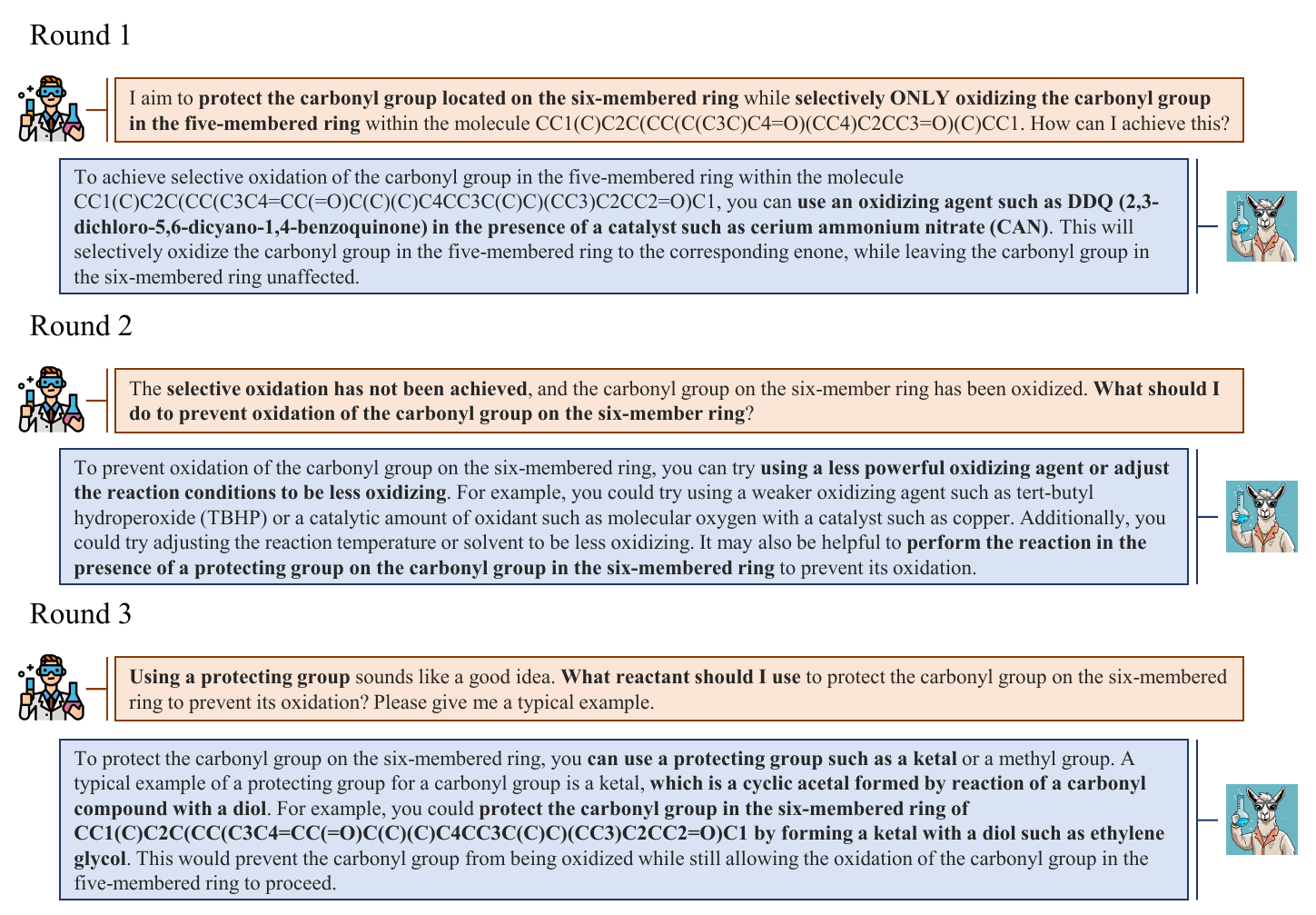}
    \caption{Example showing ChemDFM as an assistant researcher in the design of experiment through free-form dialogue. \textbf{Key points of the answer} are marked in bold. More examples can be found in Appendix Section~\ref{sec:dbc}.}
    \label{fig:multi}
\end{wrapfigure}

To function as a competent AI assistant researcher, an LLM needs not only strong chemistry skills, but also language skills to comprehend, reason, and communicate with human researchers, primarily in natural language. In the following, we test ChemDFM in two typical scenarios faced by chemistry researchers: reading papers and designing experiments, both of which demand expertise in chemical and natural language skills. 

Reading literature and other technical papers is an indispensable part of a researcher’s daily routine. Oftentimes, researchers come across new concepts or expressions that can hinder their understanding of the material. An LLM-based reading partner or assistant can provide instant explanations and answers to such questions. In Figure~\ref{fig:one}, we compare the answers generated by ChemDFM with those from other LLMs.
We have provided three example questions, with more examples in Appendix Section~\ref{sec:pr}, which are generally consistent with the analysis below.
To prevent information leakage, the questions were constructed from chemistry papers published in 2023 only. Since ChemDFM only learned from papers published before 2022, this approach ensures ChemDFM has not learned the answers during training and simulates ChemDFM's performance as a reading partner or tutor when reading new papers. \textbf{Q1} represents a question of widely known domain knowledge. All LLMs including ChemDFM provide good answers. However, when questions involve new molecules and reactions (\textbf{Q2}~\citep{yin2023total} \& \textbf{Q3}~\citep{dargo2023mesesamol}), the performances differ. Specifically, LLaMa-2 and Galactica primarily rely on retrieving knowledge from memory, which can result in numerous knowledge points that are correct but irrelevant or even misleading in the context of the questions.
GPT-4 shows a primary level of ability to answer questions based on the provided molecules and/or reactions. It effectively answers \textbf{Q2} but struggles with more complex questions involving complicated molecules such as \textbf{Q3}. In \textbf{Q3}, GPT-4 fails to fully recognize the underlying chemical aspects of the question and proposes methods that could violate the molecule's catalytic activity. It is also worth noticing that as GPT-4 is a closed-source LLM, it is uncertain whether the literature used to construct the questions is included in GPT-4's training corpus. Therefore, these "new papers" may not be new to GPT-4.
In contrast to other LLMs, ChemDFM shows the ability to integrate memory-based knowledge while considering the situation described in the questions, providing key points that are highly relevant to the question. In terms of accuracy, relevance, and overall quality of the answers, ChemDFM largely outperforms other LLMs including GPT-4, demonstrating a better understanding of molecules and reactions, especially in the example of \textbf{Q3}. Apart from presenting key points, ChemDFM also endeavors to expand on its explanation and elaborate on the mechanism of the queried reactions or the proposed solutions, although this occasionally leads to inaccurate answers, as seen in the cases of \textbf{Q1} and \textbf{Q2}. Please refer to Appendix Section~\ref{sec:pr} for a more detailed analysis. 

A knowledgeable discussion partner who is always available and patient would be invaluable for researchers, particularly in the stage of hypothesis generation and design of experiment (DOE). Figure~\ref{fig:multi} illustrates a scenario inspired by \citet{yin2023total} that showcases ChemDFM’s potential to assist researchers in free-form dialogues as an AI research partner. In this example, a human researcher aimed to selectively oxidize one of the two carbonyl groups of a molecule. The initial solution given by ChemDFM would lead to the oxidation of both carbonyl groups. However, after being alerted and challenged by the human researcher, ChemDFM acknowledged the mistake and proposed two possible strategies: using a weaker oxidation agent/condition or introducing a protecting group. Upon the researcher's decision to use a protecting group, ChemDFM provided detailed recommendations, including a feasible agent and reaction condition. Throughout the dialogue, ChemDFM exhibited promising capabilities in comprehension (Round 1), error correction (Round 2), and detailing (Round 3), showcasing its efficacy in mastering both chemical and natural language. More examples can be found in Appendix Section~\ref{sec:dbc}.

\section{Related Work}

There have been several pioneering studies focusing on leveraging LLMs to solve chemical problems. These works typically adopt one of two general strategies. The first one treats LLMs as powerful base models for multi-task training, neglecting their greatest strength in natural language understanding and reasoning~\citep{christofidellis2023unifying,fang2023molinstructions,cao2023instructmol,zheng2023shaping,kim2024large,yu2024llasmol}. Consequently, the models devised under this framework are confined to solving the specific tasks on which they were trained, losing the ability to tackle unseen tasks or conduct
free-form human-AI collaborations. The other strategy exploits LLMs' strong natural language understanding and reasoning abilities, using them directly to handle complex chemical tasks described in natural language~\citep{hatakeyama2023prompt,cao2023moformer,boiko2023autonomous,yoshikawa2023large,m2024augmenting,ruan2024accelerated}. However, most of them suffer from the fact that generalist LLMs lack an inherent understanding of chemical language and knowledge~\citep{kristiadi2024sober}. We argue that an LLM useful in chemistry must learn and reason with both general-domain knowledge and chemical knowledge. In this work, we tried to achieve this by equipping general-domain LLMs with rich chemical knowledge of task-specific chemical models and obtained promising results.

Notably, this strategy has been successfully applied to develop specialist LLMs for several other scientific domains. For example, 
Med-PaLM~\citep{singhal2023large}
and PMC-LLaMa~\citep{wu2023pmcllama} are specialized LLMs for biology and medicine.
Similarly, ChatDoctor~\citep{li2023chatdoctor} and DrugChat~\citep{liang2023drugchat} also offer LLMs specifically for the medicine field but focus on medical inquiries and drug discoveries.
Other domain-specific LLMs have endeavored
include education~\citep{dan2023educhat}, materials science~\citep{xie2023darwin}, and geography~\citep{deng2023k2}. It is worth noting that most of these works only focus on natural language. Domain-specific languages, which differ significantly from natural languages, such as SMILES in chemistry, are often overlooked.


\section{Conclusion }

In summary, this paper introduces ChemDFM, a specialist LLM that evolves from a generalist LLM through pre-training and instruction tuning using domain knowledge in chemistry. Quantitative evaluations show ChemDFM’s strong comprehension of molecular notations and reasoning capabilities for chemical knowledge, resulting in excellent performance in a wide range of chemical tasks such as molecular design and reaction analysis. In scenarios such as paper reading and experimental design, ChemDFM shows great potential in wielding chemical and natural languages to assist researchers through dialogue-based, free-form human-AI collaborations.


\begin{ack}

This work was supported by the National Science and Technology Major Project 2023ZD0120703, the China NSFC Projects (U23B2057, 62106142 and 62120106006), and the Shanghai Municipal Science and Technology Major Project (2021SHZDZX0102).


\end{ack}

\bibliographystyle{plainnat}
\bibliography{anthology,chemdfm}


\appendix




\section{Experimental Setups}

\subsection{Domain Pre-training}
\label{sec:dp}

ChemDFM is pre-trained using the popular framework {\tt Megatron-DeepSpeed}\footnote{\url{https://github.com/microsoft/Megatron-DeepSpeed?tab=readme-ov-file}} with Zero-2~\citep{10.5555/3433701.3433727} optimization technique based on LLaMa-13B~\citep{touvron2023llama}. We train ChemDFM using AdamW~\citep{loshchilov2018decoupled} with $\left(\beta_1, \beta_2\right)=\left(0.9, 0.95\right)$. During training, our model deals with $4$M tokens per batch with a maximum sequence length of $6$K. The maximum learning rate is 5e-5 under the cosine learning rate scheduler.

\subsection{Instruction Tuning}
\label{sec:it}

To fully exploit the capabilities of the pre-trained model, we employed full-parameter tuning during the instruction tuning stage. The popular framework {\tt Deepspeed-Chat}~\citep{yao2023dschat} is leveraged with the Zero-3 optimization technique. We set the learning rate to 1e-5 with a global batch size of 256. To encourage the model to focus more on responding to the requirements rather than memorizing the patterns in prompts, we performed gradient back-propagation only on the tokens of the $\mathtt{returns}$. Specifically, the loss function of our instruction tuning is
$$
\mathcal{L} = -\frac{1}{|\mathcal{D}|}\sum^{|\mathcal{D}|}_{i=1}\sum^{n_i}_{j=1}log\mathrm{P}(r_j|\mathtt{prompt}_i, r_1, r_2, ..., r_{j-1}),
$$
where $|\mathcal{D}|$ is the size of the instruction tuning dataset and $\mathtt{retunrs}_i = (r_1, r_2,...,r_{n_i})$. We train ChemDFM using AdamW with $\left(\beta_1, \beta_2\right)=\left(0.9, 0.95\right)$ and a cosine learning rate scheduler.

\section{More Details about ChemLLMBench Evaluations}
\label{sec:quantitative}

\subsection{Molecule Recognition}

\subsubsection{Task Introduction}

The name prediction tasks take advantage of the different notations of molecules, including SMILES, IUPAC name, and molecular formula, and ask the models to translate between them. Specifically, it consists of four tasks: SMILES to IUPAC name translation~(S2I), IUPAC name to SMILES translation~(I2S), SMILES to Molecular Formula translation~(S2MF), and IUPAC name to Molecular Formula translation~(I2MF). For IUPAC names and SMILES, we normalized the predictions before calculating the accuracy scores, while for molecular formulas, only exact matches are considered correct answers.

The molecule captioning tasks further require the LLMs to not only recognize what the molecule given by SMILES is but also understand the basic chemical nature of the molecule so as to generate a brief description of it. Specifically, ChemLLMBench leverages the test set of ChEBI-20~\citep{edwards-etal-2021-text2mol} for this task. To measure the performance of this task, ChemLLMBench utilizes a series of traditional captioning metrics, including BLUE, ROUGE, and METEOR.

\subsubsection{Prompt Format}

\begin{figure}
    \centering
    \includegraphics[width=\linewidth]{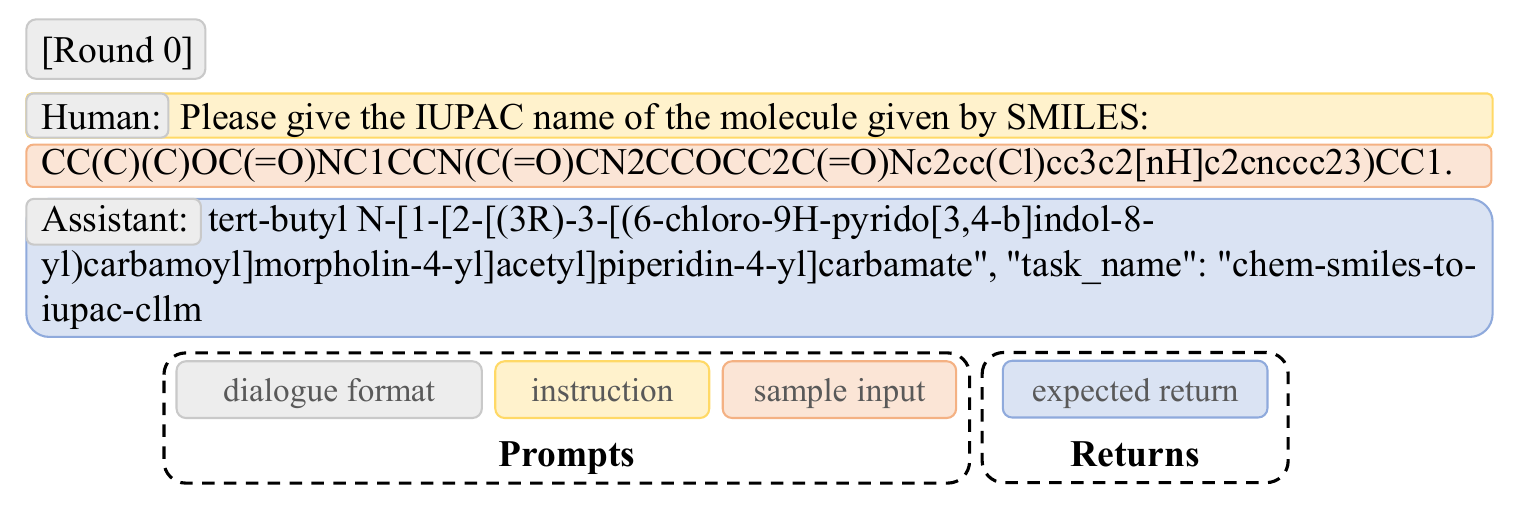}
    \caption{Prompt format of the name prediction tasks}
    \label{fig:mrpf}
\end{figure}

For the name prediction tasks, we use a simpler prompt compared with that introduced in \citet{guo2023large}. An example is shown in Figure~\ref{fig:mrpf}

For the molecule captioning task, we use the same prompt introduced in \citet{guo2023large}.

\subsection{Text-Based Molecule Design}

\subsubsection{Task Introduction}

The test set of ChEBI-20 is also exploited for this task in ChemLLMBench. Models are asked to predict the SMILES of the molecule that fits the given description. Two kinds of metrics are utilized to measure the performance of this task. The first set of metrics measures the text-based similarity of the predicted SMILES compared to the golden SMILES, which includes exact match, BLUE, and Levenshtein distance. The second set of metrics measures the chemical similarity of the predicted molecules compared to the golden molecules. That is mainly composed of the validity of the predicted SMILES and the FTS~(fingerprint Tanimoto Similarity)~\citep{tanimoto1958elementary} in terms of MACCS~\citep{maccs}, RDK\footnote{\url{https://www.rdkit.org/}}, Morgan~\citep{morgan}.

\subsubsection{Prompt Format}

We use the same prompt introduced in \citet{guo2023large}.

\subsubsection{Additional Results}

\begin{table*}[t]
    \centering
    \caption{Benchmark full test-set evaluation results of different models in text-based molecule design tasks. The best results among specialist and generalist models are highlighted in bold, respectively. \dag: reproducing results.}
    \label{tab:full}
    \resizebox{\textwidth}{!}{\begin{tabular}{lccccccc}
    \toprule
    Model & Exact$\uparrow$ & BLUE$\uparrow$ & Dis$\downarrow$ & Validity$\uparrow$ & MACCS$\uparrow$ & RDK$\uparrow$ & Morgan$\uparrow$ \\
    \midrule
    \rowcolor{grey}\multicolumn{8}{c}{\textit{task-specific specialist models}} \\
    MolXPT~\citep{liu-etal-2023-molxpt} & 21.5 & - & - & 98.3 & 0.859 & 0.757 & 0.667 \\
    Text+Chem T5~\citep{christofidellis2023unifying} & \textbf{32.2} & \textbf{0.853} & \textbf{16.87} & 94.3 & \textbf{0.901} & \textbf{0.816} & \textbf{0.757} \\
    Mol-Instruction~\citep{fang2023molinstructions} & 0.2 & 0.345 & 41.4 & \textbf{100} & 0.412 & 0.231 & 0.147 \\
    \midrule
    \rowcolor{grey}\multicolumn{8}{c}{\textit{LLM-based generalist models}} \\
    Galactica-30B~(10-shot)\textsuperscript{\dag} & 0.3 & 0.295 & 64.3 & 82.2 & 0.356 & 0.239 & 0.186 \\
    \textbf{ChemDFM-13B} & \textbf{43.2} & \textbf{0.839} & \textbf{16.9} & \textbf{97.6} & \textbf{0.901} & \textbf{0.829} & \textbf{0.759} \\
    \bottomrule
    \end{tabular}}
\end{table*}

To achieve a fair comparison with task-specific specialist models, we evaluate the performance of ChemDFM on the full test set of ChEBI-20 on this task. The results are illustrated in Table~\ref{tab:full}. ChemDFM surpasses the performance of the advanced specialist models on the major metrics while achieving comparable performance on others. Specifically, ChemDFM outperforms the specialist models on exact match scores and all three FTS-based similarity scores, which indicates that ChemDFM can make more reliable predictions based on the descriptions compared with specialist models.

\subsection{Molecular Property Prediction}

\subsubsection{Task Introduction}

The molecular property prediction tasks in ChemLLMBench consist of five tasks from MoleculeNet benchmark~\citep{wu2018moleculenet}, including BACE, BBBP, HIV, ClinTox, and Tox21. Among them, BACE and BBBP are each a balanced binary classification task. HIV is an unbalanced binary classification task. ClinTox and Tox21 comprise two and twenty-one unbalanced binary classification tasks, respectively.

\subsubsection{Prompt Format}

We use the same prompts introduced in \citet{guo2023large}.

\subsubsection{Additional Results}

\begin{table*}[t]
    \centering
    \caption{AUC-ROC scores~\citep{bradley1997use} of different models under different settings in molecular property prediction tasks. \dag reproducing results (The results of GPT-4 were obtained in January 2024).}
    \label{tab:fewshot}
    \begin{tabular}{lccccc}
    \toprule
    Model & BACE$\uparrow$ & BBBP$\uparrow$ & ClinTox$\uparrow$ & HIV$\uparrow$ & Tox21$\uparrow$ \\
    \midrule
    \rowcolor{grey}\multicolumn{6}{c}{\textit{LLM-based generalist models}} \\
    GPT-4~(0-shot)\textsuperscript{\dag} & 62.5 & 61.5 & 51.6 & 65.9 & 55.2 \\
    GPT-4~(8-shot)\textsuperscript{\dag} & 45.9 & 61.8 & 59.3 & 50.8 & 60.6 \\
    LLaMa-2-13B-chat~(0-shot)\textsuperscript{\dag} & 26.0 & 60.3 & 45.7 & 29.0 & 51.7 \\
    LLaMa-2-13B-chat~(8-shot)\textsuperscript{\dag} & 72.9 & 52.3 & 42.1 & 70.8 & 45.9 \\
    Galactica-30B~\citep{taylor2022galactica} & 72.7 & 59.6 & 82.2 & \textbf{75.9} & 68.5 \\
    \textbf{ChemDFM-13B~(0-shot)} & \underline{78.4} & \underline{66.7} & \textbf{89.9} & \underline{73.6} & \textbf{79.8} \\
    \textbf{ChemDFM-13B~(8-shot)} & \textbf{81.7} & \textbf{67.9} & \underline{85.3} & 73.3 & \underline{76.7} \\
    \bottomrule
    \end{tabular}
\end{table*}

During evaluations, we leverage a popular and more challenging dataset split provided by DeepChem library~\citep{deepchem}. We reproduce the results of the baseline models, including GPT-4, LLaMa-2-13B-chat, and Galactica~(30B). Apart from the results in the Quantitative Evaluation Section of the main text, we also conduct few-shot experiments. The results are shown in Table~\ref{tab:fewshot}. It is worth noticing that the performances under the few-shot setting are not always better than those under the zero-shot setting. That may be a result of the scaffold-vertical dataset split we use in our experiments. Because under the scaffold-vertical setting, the exemplars provided by the training split may be less helpful for the test samples.

\subsection{Reaction Prediction and Retrosynthesis}

\subsubsection{Task Introduction}

In ChemLLMBench, there are four types of tasks targeted at evaluating models' capabilities of reaction understanding. The yield prediction tasks ask models to predict whether the given reaction is a high-yield reaction and are constructed based on two High-Throughput experimentation (HTE) datasets: the Buchwald-Hartwig dataset~\citep{ahneman2018predicting} and the Suzuki-Miyaura dataset~\citep{reizman2016suzuki}. The reaction prediction task asks the model to predict the product of the given reaction. ChemLLMBench utilizes the USPTO-MIT dataset~\citep{jin2017predicting} for this task. The reagent selection tasks focus on selecting the reagent that can maximize the yield of the reaction from a list of candidates. ChemLLMBench constructs three reagent selection tasks based on the dataset proposed by \citet{perera2018platform}. The retrosynthesis task focuses on predicting the reactants of the given reactions and is constructed based on the USPTO-50K dataset~\citep{schneider2016s}. Accuracy is utilized to measure the performances except for the ligand selection task which uses top 50\% accuracy.

\subsubsection{Prompt Format}





\begin{figure*}
    \centering
    \includegraphics[width=\linewidth]{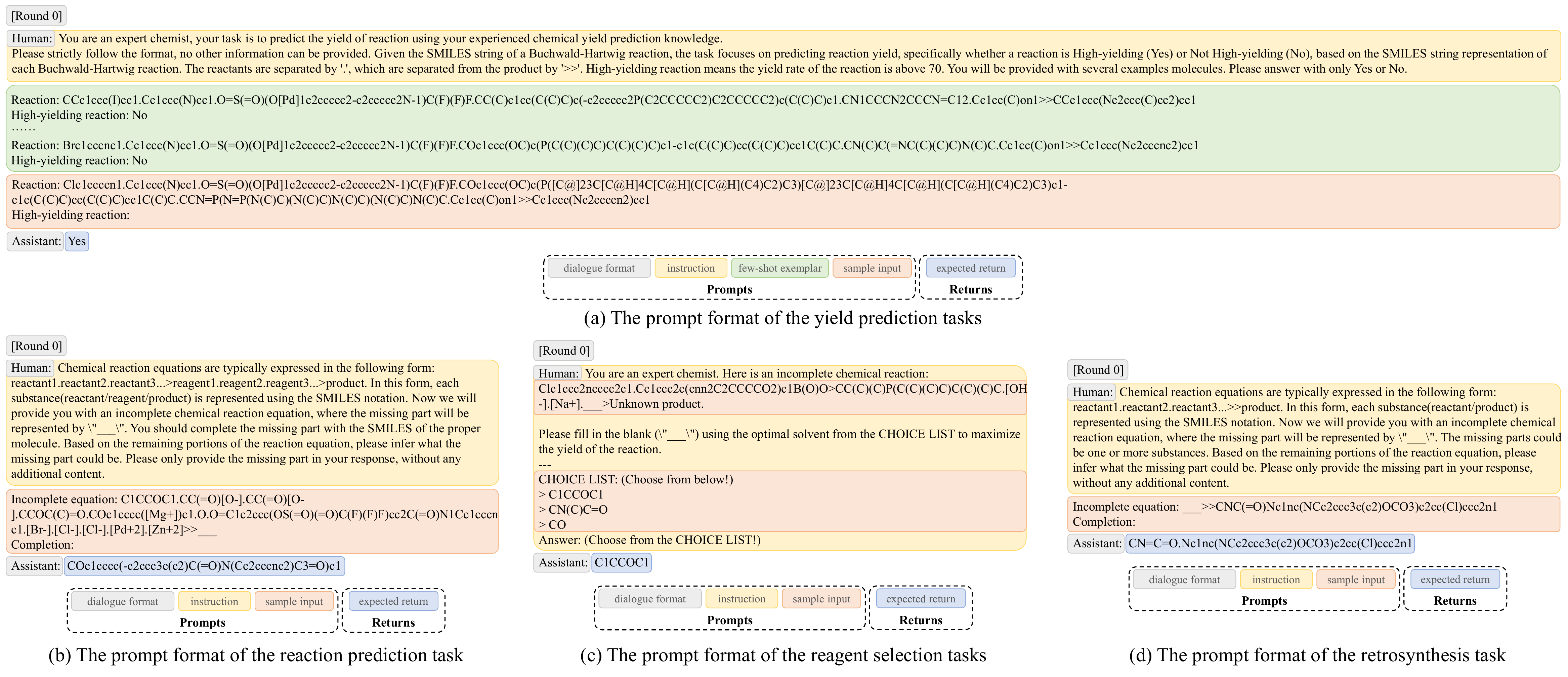}
    \caption{Prompt format of the reaction prediction and retrosynthesis tasks}
    \label{fig:prompt}
\end{figure*}

We reformat the prompt provided by \citet{guo2023large} using the SMILES notations for reactions. Specifically, the examples of our prompts are illustrated in Figure~\ref{fig:prompt}.

\subsubsection{Additional Results}

\begin{table}[t]
    \centering
    \caption{Accuracy scores of different models in yield prediction tasks. B-H and Suzuki stand for the Buchwald-Hartwig dataset and the Suzuki-Miyaura dataset, respectively. \dag: results from \citet{guo2023large}. *: 8-shot results}
    \label{tab:yp}
    \begin{tabular}{lcc}
    \toprule
    Model & B-H & Suzuki \\
    \midrule
    \rowcolor{grey}\multicolumn{3}{c}{\textit{task-specific specialist models}} \\
    UAGNN~\citep{kwon2022uncertainty} & 96.5 & 95.7 \\
    \midrule
    \rowcolor{grey}\multicolumn{3}{c}{\textit{LLM-based generalist models}} \\
    GPT-4\textsuperscript{\dag}\textsuperscript{*} & \underline{80.0} & \underline{76.4} \\
    LLaMa-2-13B-chat\textsuperscript{\dag} & 0.8 & 0.6 \\
    Galactica-30B\textsuperscript{\dag} & 0.0 & 0.8 \\
    \textbf{ChemDFM-13B} & \textbf{82.7} & \textbf{79.3} \\
    \bottomrule
    \end{tabular}
\end{table}
\begin{table}[t]
    \centering
    \caption{Benchmark results of different models in reaction prediction tasks. \dag: results from \citet{guo2023large}.}
    \label{tab:rp}
    \begin{tabular}{lcc}
    \toprule
    Model & Accuracy & Validity \\
    \midrule
    \rowcolor{grey}\multicolumn{3}{c}{\textit{task-specific specialist models}} \\
    Chemformer~\citep{irwin2022chemformer} & 93.8 & 100 \\
    Mol-Instruction~\citep{fang2023molinstructions} & 4.5 & 100 \\
    InstructMol~\citep{cao2023instructmol} & 53.6 & 100 \\
    \midrule
    \rowcolor{grey}\multicolumn{3}{c}{\textit{LLM-based generalist models}} \\
    GPT-4 (20-shot)\textsuperscript{\dag} & \underline{23.0} & 93.0 \\
    LLaMa-2-13B-chat (20-shot)\textsuperscript{\dag} & 3.2 & 72.2 \\
    Galactica-30B (5-shot)\textsuperscript{\dag} & 3.6 & \underline{94.8} \\
    \textbf{ChemDFM-13B (0-shot)} & \textbf{49.0} & \textbf{98.0} \\
    \bottomrule
    \end{tabular}
\end{table}
\begin{table}[t]
    \centering
    \caption{Benchmark results of different models in retrosynthesis tasks. \dag: results from \citet{guo2023large}.}
    \label{tab:retro}
    \begin{tabular}{lcc}
    \toprule
    Model & Accuracy & Validity \\
    \midrule
    \rowcolor{grey}\multicolumn{3}{c}{\textit{task-specific specialist models}} \\
    Chemformer~\citep{irwin2022chemformer} & 53.6 & 100 \\
    \midrule
    \rowcolor{grey}\multicolumn{3}{c}{\textit{LLM-based generalist models}} \\
    GPT-4 (5-shot)\textsuperscript{\dag} & \underline{11.4} & 89.0 \\
    LLaMa-2-13B-chat (20-shot)\textsuperscript{\dag} & 0.0 & 72.8 \\
    Galactica-30B (5-shot)\textsuperscript{\dag} & 1.6 & \textbf{94.8} \\
    \textbf{ChemDFM-13B (0-shot)} & \textbf{12.0} & \underline{91.0} \\
    \bottomrule
    \end{tabular}
\end{table}
\begin{table}[t]
    \centering
    \caption{Benchmark results of different models in reagent selection tasks. We report the result in accuracy scores except for Ligand Selection where we report the top 50\% accuracy score. \dag: results from \citet{guo2023large}.}
    \label{tab:rs}
    \begin{tabular}{lccc}
    \toprule
    Model & Reactant & Solvent & Ligand \\
    \midrule
    \rowcolor{grey}\multicolumn{4}{c}{\textit{LLM-based generalist models}} \\
    GPT-4\textsuperscript{\dag} & \textbf{29.9} & \textbf{52.6} & \textbf{53.4}  \\
    LLaMa-2-13B-chat\textsuperscript{\dag} & 14.5 & 5.0 & 28.4 \\
    Galactica-30B\textsuperscript{\dag} & 10.7 & 10.4 & 3.0 \\
    \textbf{ChemDFM-13B} & \underline{24.0} & \underline{12.0} & \underline{35.0} \\
    \bottomrule
    \end{tabular}
\end{table}

The complete results for the yield prediction tasks, the reaction prediction task, the reagent selection tasks, and the retrosynthesis tasks are shown in Table~\ref{tab:yp}, Table~\ref{tab:rp}, Table~\ref{tab:rs}, and Table~\ref{tab:retro}, respectively.

\section{More Qualitative Analysis}
\label{sec:qualitative}

\subsection{Paper Reading}
\label{sec:pr}

\begin{figure*}
    \centering
    \includegraphics[width=\linewidth]{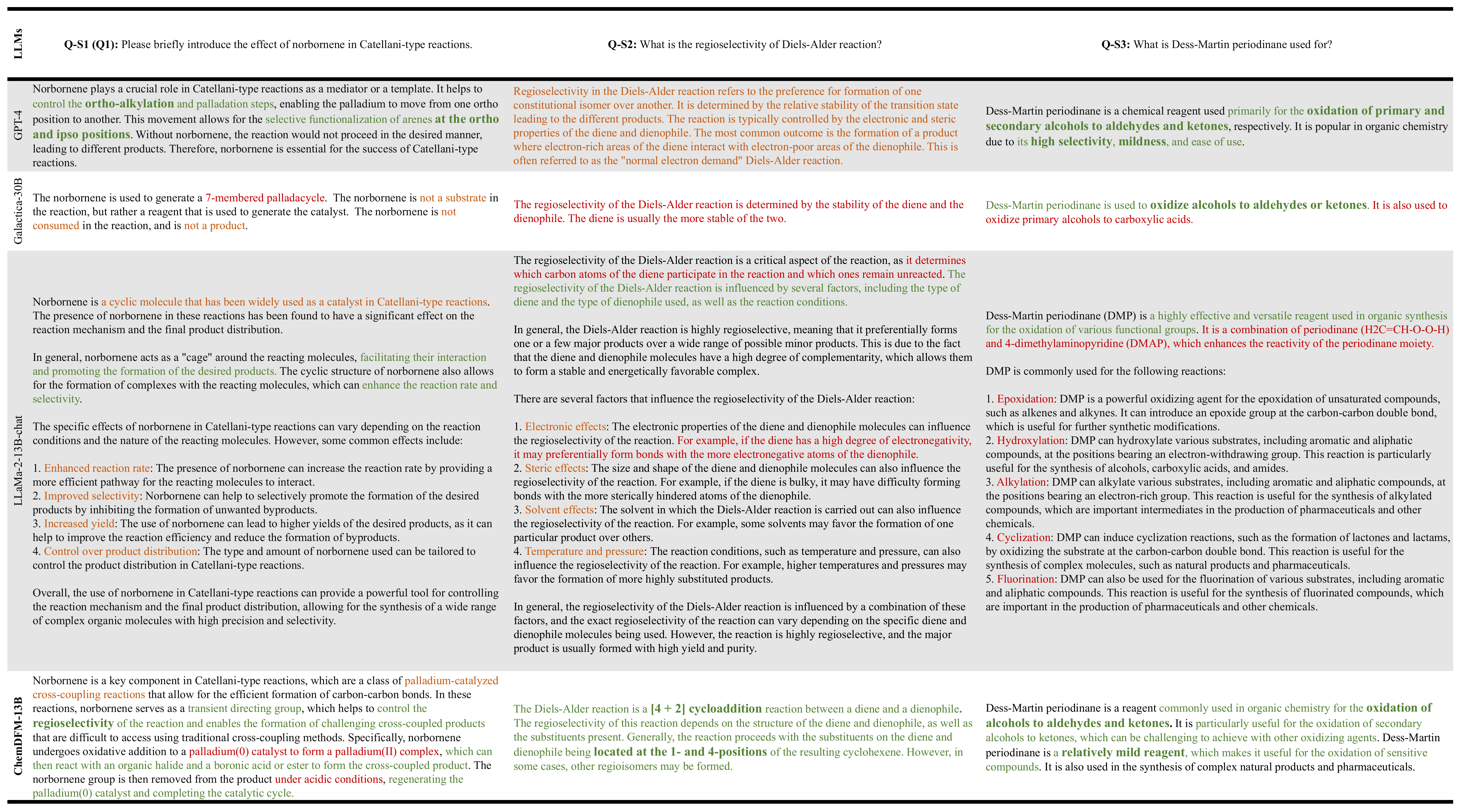}
    \caption{Examples of paper reading where only widely known domain knowledge is involved. \textcolor{green}{correct and relevant information} in the replies is marked in green, \textcolor{yellow}{correct but irrelevant information} in yellow, and \textcolor{red}{wrong information} in red. \textbf{Key points of the answer} are marked in bold.}
    \label{fig:1-1}
\end{figure*}
\begin{figure*}
    \centering
    \includegraphics[width=\linewidth]{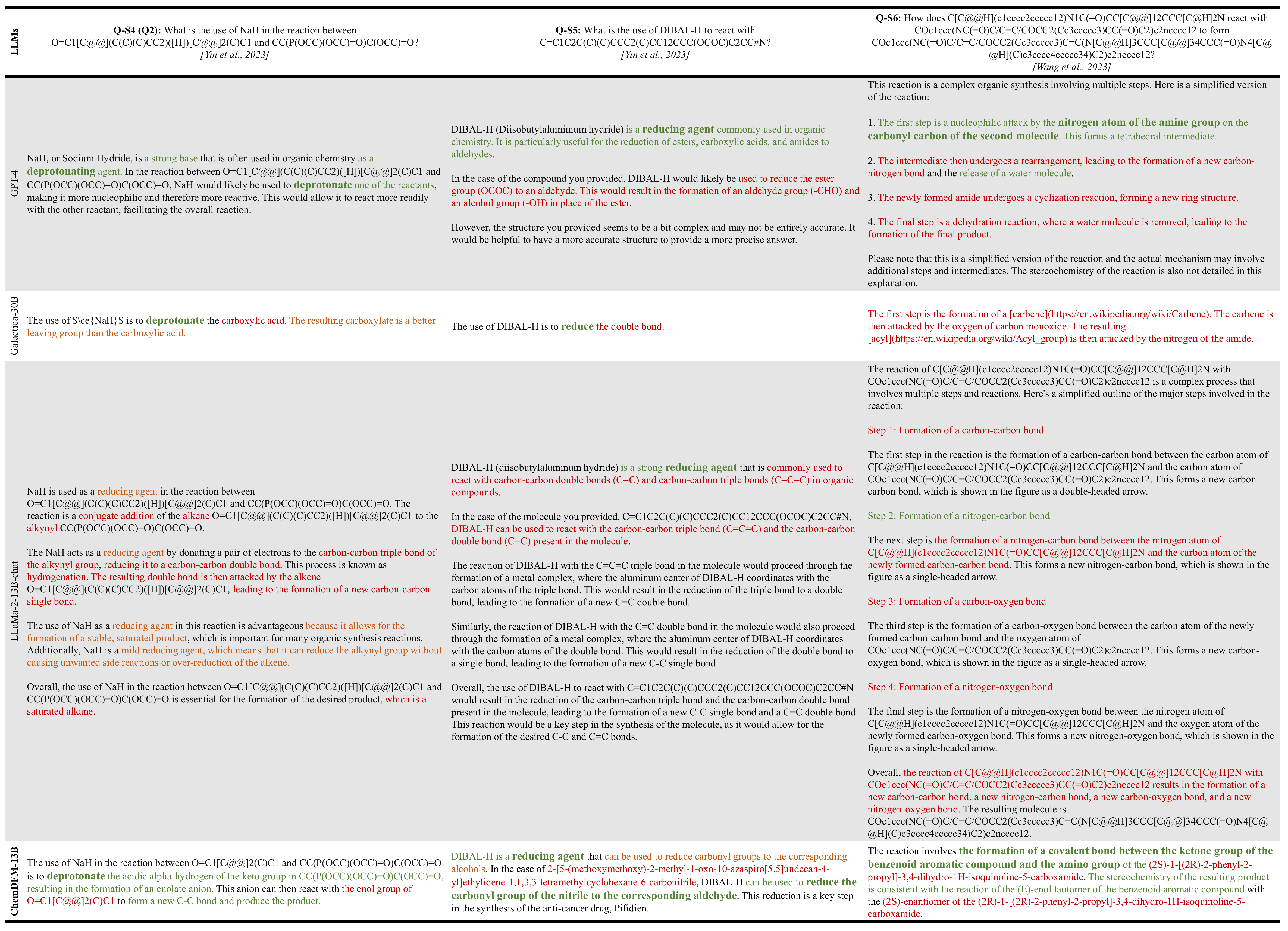}
    \caption{Examples of paper reading where new molecules and reactions are involved. \textcolor{green}{correct and relevant information} in the replies is marked in green, \textcolor{yellow}{correct but irrelevant information} in yellow, and \textcolor{red}{wrong information} in red. \textbf{Key points of the answer} are marked in bold.}
    \label{fig:1-2}
\end{figure*}
\begin{figure*}
    \centering
    \includegraphics[width=\linewidth]{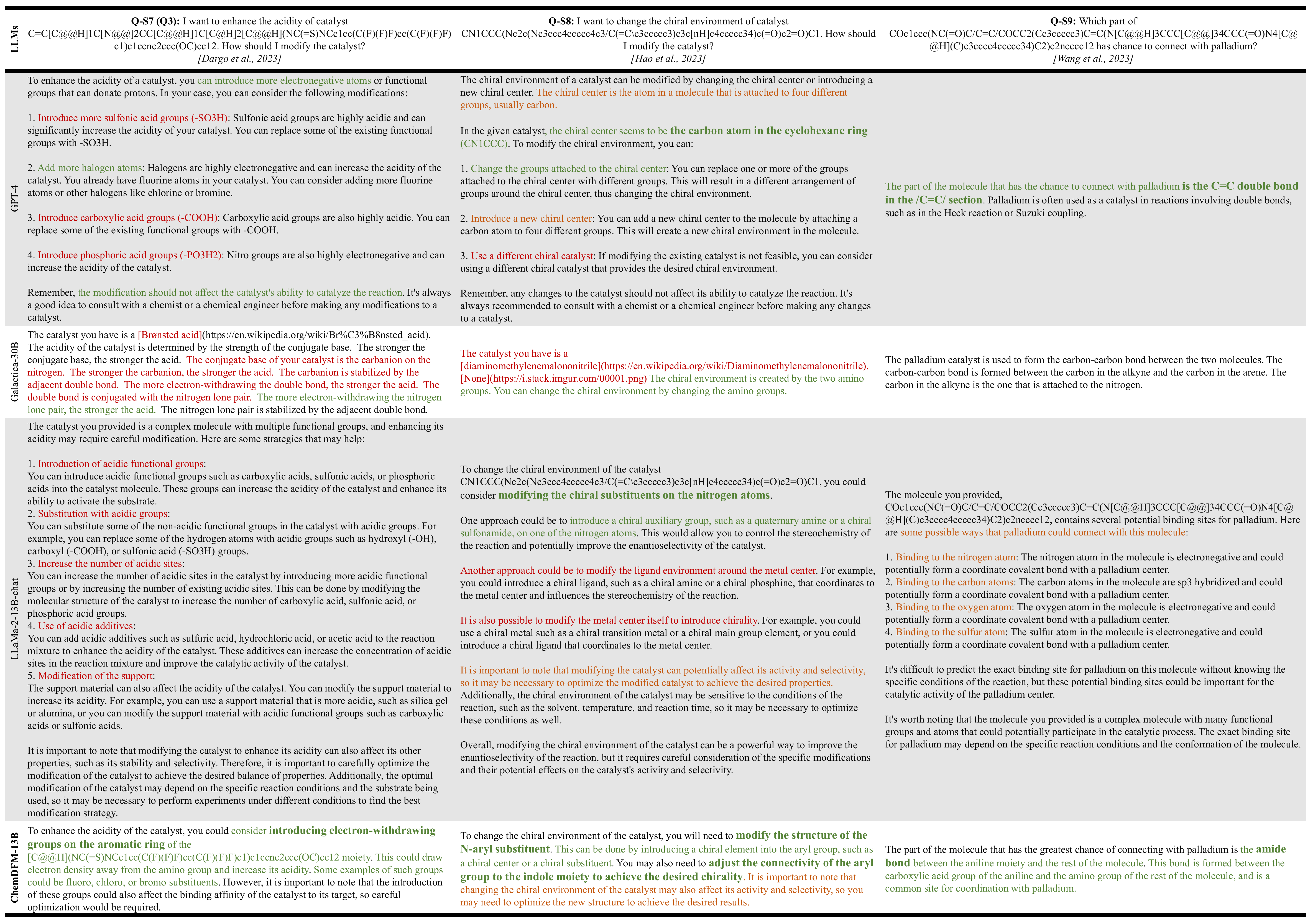}
    \caption{Examples of paper reading where new molecules and reactions are involved. \textcolor{green}{correct and relevant information} in the replies is marked in green, \textcolor{yellow}{correct but irrelevant information} in yellow, and \textcolor{red}{wrong information} in red. \textbf{Key points of the answer} are marked in bold.}
    \label{fig:1-3}
\end{figure*}

We first test the models with questions that only involve known knowledge~(Figure~\ref{fig:1-1}).

\textbf{Q-S1 (Q1)} is an example of knowledge-intense questions. Models only need to memorize the details and mechanisms of Catellani-type reactions~\citep{catellani1997complex} to answer the question correctly. The key point of the answer to this question is ``\textbf{regioselectivity}''. While Galactica can hardly answer the question and LLaMa-2 misses the key point of the answer, ChemDFM accurately captures the key point to answer the question and provides a comprehensive answer. GPT-4 gives the best reply as it not only points out ``regioselectivity'' but also gives the result of the regioselectivity of norbornene. ChemDFM is the only model that tries to provide a detailed description of the mechanism behind the reaction. However, it makes minor mistakes when doing so.

\textbf{Q-S2} asks for the regioselectivity of the Diels-Alder reaction~\citep{kloetzel1948diels}. Only ChemDFM successfully answers the key points to this question, which is the result of the regioselectivity. GPT-4 provides a detailed introduction to the Diels-Alder reaction and regioselectivity but fails to answer the specific regioselectivity of the Diels-Alder reaction, while LLaMa-2 only gives the factors that could influence the regioselectivity. They do not answer the question.

As for \textbf{Q-S3}, ChemDFM, Galactica, and GPT-4, all capture the key point to the answer (``\textbf{the oxidation of alcohols to aldehydes and ketones}''), while ChemDFM and GPT-4 further answer more properties of the Dess-Martin periodinane~\citep{dess1983readily}. LLaMa-2, on the other hand, gives numerous wrong arguments and misses the key points.

Then, we ask the models about new molecules and new reactions which are published after January 2022. In this way, we can ensure minimal risk of data leakage and evaluate the models' capability to handle unforeseen situations. The results are shown in Figure~\ref{fig:1-2} and Figure~\ref{fig:1-3}.

\textbf{Q-S4 (Q2)} is constructed based on \citet{yin2023total}. Because the reaction mentioned in the question is a novel instance, models need to correctly identify the reaction and discover the mechanisms of it before answering the question. In practice, Galactica successfully identifies the key point of the answer, ``\textbf{deprotonate}'', but fails to provide other useful information. LLaMa-2, in its reply, fails to identify the reaction mentioned in the question. Most of the information about {\tt NaH} in its reply is correct but irrelevant to the reaction. GPT-4 identifies the key point of the answer but only gives a rough description of the mechanism of how it works. ChemDFM not only correctly identifies the key point of the answer but also provides an almost correct description of the mechanism.

\textbf{Q-S5} is also constructed based on \citet{yin2023total}. All the models can recognize the DIBAL-H as a reducing agent, which is existing knowledge. However, only ChemDFM successfully identifies the reaction site of the new molecule, indicating its strong capabilities to handle unforeseen situations where new molecules and reactions are involved. The main mistake that ChemDFM makes is providing the wrong IUPAC name, which is a challenging task for LLMs even as a separate task~(see Table~2 in the main text).

\textbf{Q-S6} is constructed based on \citet{wang2023asymmetric} and asks directly for the mechanism of the given reaction. Among the answers, the answer of ChemDFM is the most precise. Galactica and LLaMa-2 give nearly no correct information. Although GPT-4's answer contains the correct reaction process, it also contains auxiliary processes that do not happen during the reaction, which masks the whole mechanism predicted by GPT-4 wrong. ChemDFM answers the correct reaction process with no excess. The only mistakes ChemDFM makes are again providing the wrong IUPAC names, which is a challenging task for LLMs even as a separate task~(see Table~2 in the main text).

We also ask several questions focusing more on molecules and less on reactions.

\textbf{Q-S7 (Q3)}, constructed based on \citet{dargo2023mesesamol}, focus on the modification of catalyst molecules. The molecule mentioned in the question is a novel instance and models need to infer the chemical properties of that molecule to answer the question. The key point of the answer is ``\textbf{introducing electron-withdrawing groups on the aromatic rings}'' as this method has the potential to increase the acidity while keeping the catalytic ability of the molecule. Among the LLMs, only ChemDFM successfully answers the key point, while others either fail to provide any specific solutions or give wrong solutions which will damage the catalytic ability of the molecule.

\textbf{Q-S8}, constructed based on \citet{hao2023axially}, focus on the modification of chiral environment. In the given molecule, there are two chiral centers. GPT-4 identifies the point chiral center and provides candidate methods that are not all correct. The other three models identify the axial chirality which is challenging to identify with only the SMILES notation. Among the three models, Galactica gives almost no detailed method to change the chiral environment, LLaMa-2 gives one correct method with more wrong ones, and ChemDFM provides two correct methods one of which is targeting specifically the axial chirality.

\textbf{Q-S9} is constructed on \citet{wang2023asymmetric}. It asks for the coordinate sites between the given molecule and palladium. There are a total of three coordinate sites. GPT-4 and ChemDFM each identify one of them, while Galactica and LLaMa-2 fail to identify any.

\subsection{Dialogue-Based Human-AI Collaboration}
\label{sec:dbc}

We demonstrate two more examples of dialogue-based human-AI collaboration based on ChemDFM here. The dialogues are also inspired by the recently published papers to minimize the risk of data leakage and evaluate ChemDFM's capability to handle unforeseen situations during human-AI collaboration.

\begin{figure*}
    \centering
    \includegraphics[width=\linewidth]{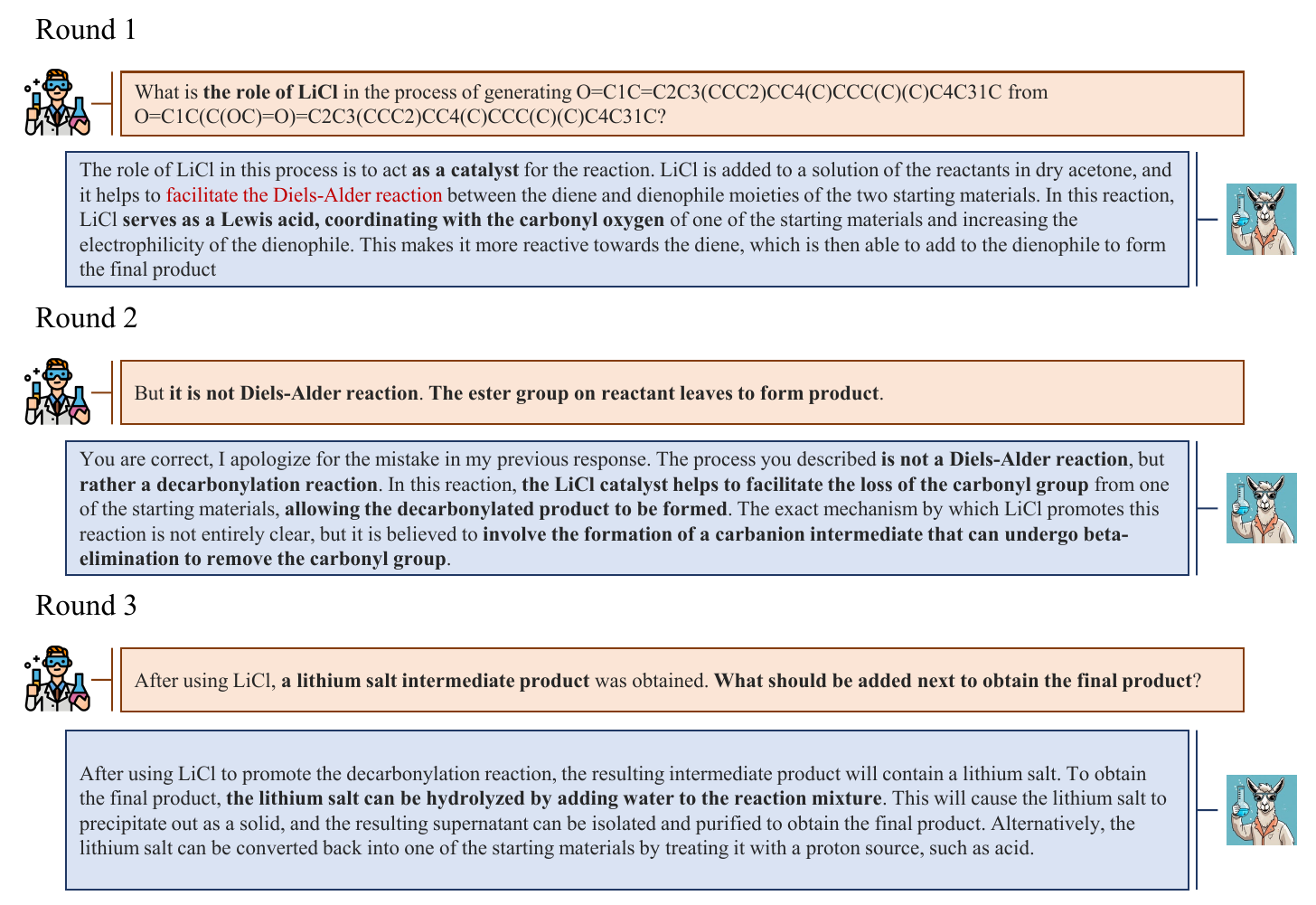}
    \caption{Example showing ChemDFM as an assistant researcher in the design of experiment through free-form dialogue. \textbf{Key points of the answer} are marked in bold and \textcolor{red}{wrong information} in red. The dialogue is inspired by \citet{yin2023total}.}
    \label{fig:2-1}
\end{figure*}
\begin{figure*}
    \centering
    \includegraphics[width=\linewidth]{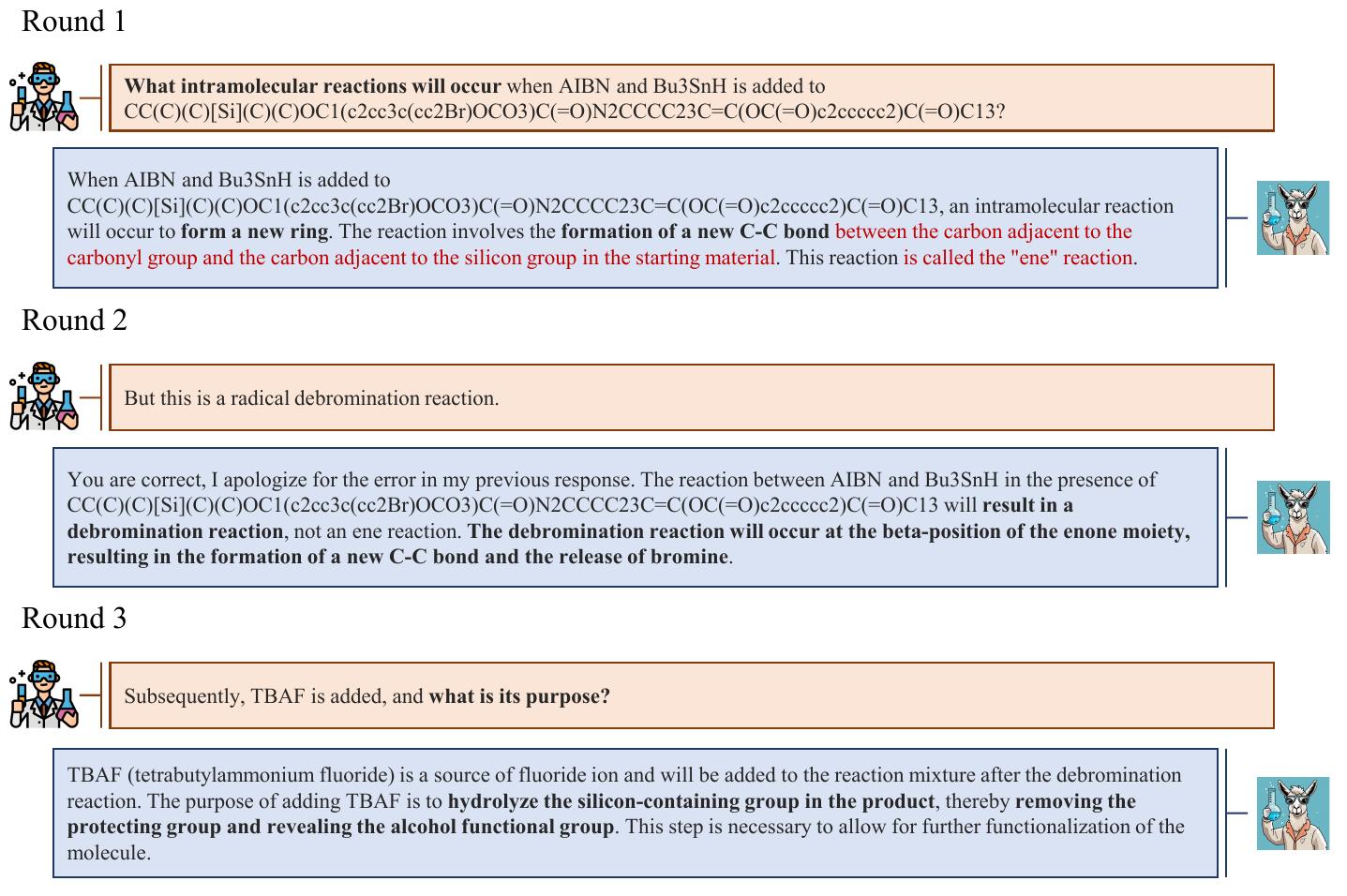}
    \caption{Example showing ChemDFM as an assistant researcher in the design of experiment through free-form dialogue. \textbf{Key points of the answer} are marked in bold and \textcolor{red}{wrong information} in red. The dialogue is inspired by \citet{zhuang2023catalytic}.}
    \label{fig:2-2}
\end{figure*}

The dialogue shown in Figure~\ref{fig:2-1} is inspired by \citet{yin2023total}. During the dialogue, the human researcher first asks for the role of {\tt LiCl} in the given reaction. ChemDFM successfully identifies the {\tt LiCl} as a catalyst while misjudging the type of the reaction. To correct the answer, the human researcher points out the key error in the answer with some important details of the reaction (which can be easily discovered by comparing the product with the reactant). ChemDFM then corrects its mistake with even more details about the reaction process. To further validate whether ChemDFM fully understands the unforeseen reaction, we continue to ask about the post-processing procedure which is necessary to get the final product. ChemDFM successfully captures the key point to the question and gives the correct answer.

The dialogue shown in Figure~\ref{fig:2-2} is inspired by \citet{zhuang2023catalytic}. ChemDFM first gives a partially correct answer to the question from the human researcher where it misjudges the position of the newly formed {\tt C-C} bond and the type of the reaction. With the help of human correction, ChemDFM then realizes the mistakes and corrects them. Then the human researcher further asks about the next reaction that is conducted in \citet{zhuang2023catalytic} without clarifying the current molecule composition of the system or restating the previous reaction. ChemDFM can infer this information from the dialogue history and correctly answer the question.

In these dialogues, ChemDFM shows promising capabilities in handling unforeseen situations, error correction, and inferring information from dialogue history.
These capabilities can be attributed to the fact that ChemDFM comprehends both natural language and chemical language. This allows a universal language protocol established between ChemDFM and human researchers, enabling meaningful human-AI collaborations.


\end{document}